\documentclass{article}
\usepackage[english]{babel}
\usepackage{subcaption}
\usepackage{fontawesome5}
\PassOptionsToPackage{table}{xcolor}
\usepackage[final]{colm2026_conference}

\usepackage{microtype}
\usepackage{hyperref}
\usepackage{url}
\usepackage{booktabs}
\usepackage{lineno}
\usepackage{caption}

\usepackage{xcolor}  
\usepackage{amsfonts}
\usepackage{listings}
\usepackage{float}
\usepackage{minted}
\usepackage[T1]{fontenc}
\usepackage[utf8]{inputenc}
\usepackage{inconsolata}
\usepackage{placeins}
\usepackage{multirow}
\usepackage{wrapfig}
\usepackage{graphicx}
\usepackage{xspace}
\usepackage{amsmath}
\usepackage[ruled,vlined]{algorithm2e}
\usepackage{siunitx}
\usepackage{booktabs}
\usepackage{pifont}
\usepackage{threeparttable}

\newcommand{\xmark}{\ding{55}}

\definecolor{darkblue}{rgb}{0, 0, 0.5}
\hypersetup{colorlinks=true, citecolor=darkblue, linkcolor=darkblue, urlcolor=darkblue}

\newcommand{\modelname}{\textsc{ArtifactLinker}\xspace}
\newcommand{\benchname}{\textsc{ArtifactBench}\xspace}

\definecolor{bestgray}{gray}{0.8}
\definecolor{graphrowcolor}{gray}{0.93}
\newcommand{\huggingface}{\raisebox{-1.5pt}{\includegraphics[height=1.05em]{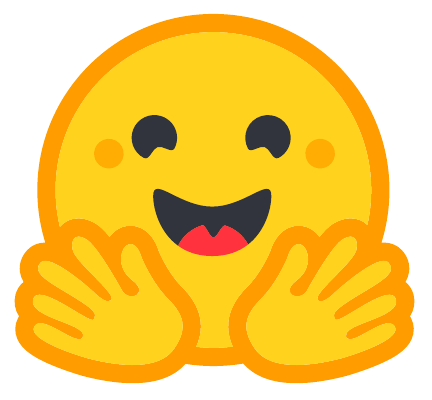}}\xspace}
\newcommand{\github}{\raisebox{-1.5pt}{\includegraphics[height=1.05em]{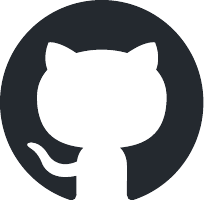}}\xspace}

\title{ArtifactLinker: Linking Scientific Artifacts for \\Automatic State-of-the-Art Discovery}

\author{Haofei Yu\textsuperscript{1,2}\thanks{Work done during the internship at the Allen Institute for AI.}\quad
Jiaxuan You\textsuperscript{1}\quad
Peter Clark\textsuperscript{2}\quad \\
\textbf{Bodhisattwa Prasad Majumder}\textsuperscript{\textbf{2}}\quad
\textbf{Kyle Richardson}\textsuperscript{\textbf{2}} \\
\textsuperscript{1}University of Illinois Urbana-Champaign \quad
\textsuperscript{2}Allen Institute for AI \\
\texttt{haofeiy2@illinois.edu, kyler@allenai.org}
}

\begin{document}

\ifcolmsubmission
\linenumbers
\fi

\maketitle

\vspace{-1.5em}
\begin{flushleft}
\hspace{2mm}\github~\texttt{Code: \href{https://github.com/allenai/artifact-linker}{https://github.com/allenai/artifact-linker}}\\ \vspace{2mm}
\hspace{2mm}\huggingface~\texttt{Data: \href{https://huggingface.co/datasets/lwaekfjlk/artifact-bench}{https://huggingface.co/datasets/lwaekfjlk/artifact-bench}}
\end{flushleft}
\vspace{1.5em}

\begin{abstract}
Scientific artifacts such as models and datasets are foundations for research. With the rapid growth of platforms like HuggingFace, researchers now have access to a large number of artifacts. Yet, a key challenge remains: how can we automatically discover the state-of-the-art (SOTA) model for a given dataset by fully leveraging existing artifacts? We formalize this task as \textbf{automatic SOTA discovery} by modeling HuggingFace as an artifact graph, where nodes are models/datasets and edges represent evaluations. We propose \modelname, a two-stage framework: (1) ranking promising unobserved model--dataset links using Graph Neural Networks (GNNs) or graph-augmented Large Language Models (LLMs), and (2) verifying top-ranked links via coding experiments with LLM-based agents. We further introduce a benchmark named \benchname with 14,053 artifacts and 51,337 relations to evaluate the performance of both stages. Results show that (1) graph structures between existing artifacts are effective for missing link prediction; (2) end-to-end ranking and verification with \modelname help discover potential SOTA results and research insights.
\end{abstract}

\section{Introduction}
\vspace{-1mm}

Scientific artifacts are fundamental building blocks of modern research~\citep{Heumller2020PublishOPA,Cooper2022ASRA,Johnson2019ArtifactBasedRHA}. Models and datasets on the HuggingFace Hub are representative examples: researchers routinely share, reuse, and extend these artifacts to support reproducibility and drive progress~\citep{Mari2023APWA,Lissa2020WORCSAWA}. As the machine learning community continues to produce an increasing number of artifacts across sub-domains~\citep{Castao2024HowDMA,Ait2023OnTSA,Laufer2025AnatomyOAA}, a natural question arises: \textit{Can we leverage existing artifacts to support automatic scientific discovery?} We study the HuggingFace ecosystem as a concrete setting for this question because it provides a large, active, and executable collection of models, datasets, and libraries.

\begin{figure}[t]
\centering
\begin{minipage}[t]{\textwidth}
\vspace{0pt}
\centering
\includegraphics[width=\textwidth]{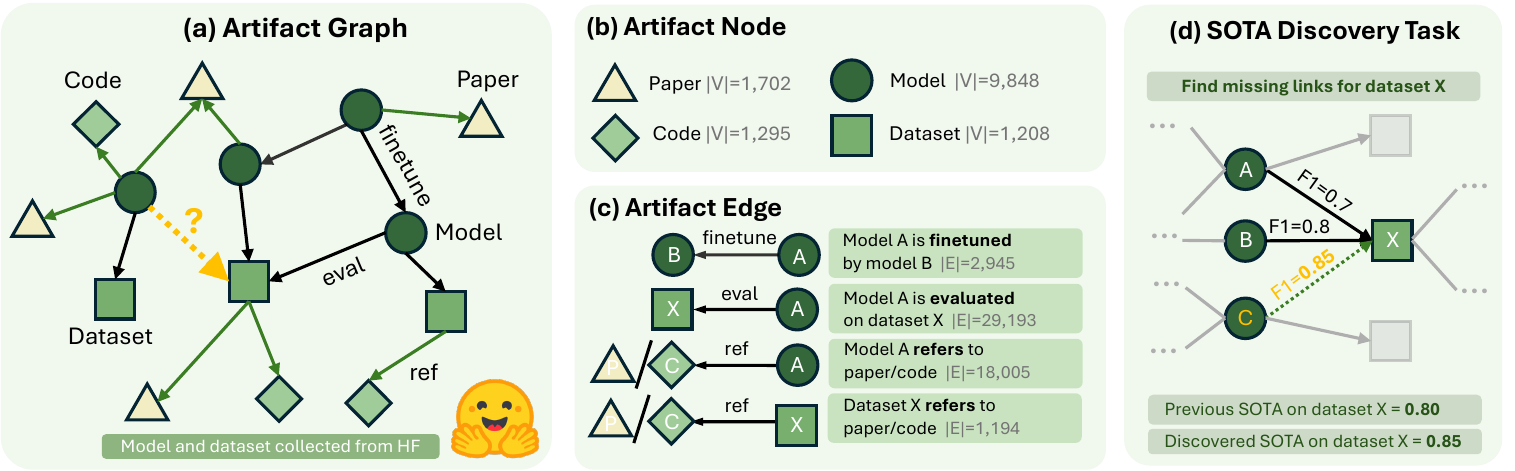}
\caption{\textbf{Artifact graph structure and SOTA discovery task formulation.}
\textbf{(a) Example graph.} Visualization of the artifact graph, highlighting its sparsity and missing links across artifact types.
\textbf{(b) Node statistics.} Distribution of nodes across artifact categories.
\textbf{(c) Edge statistics.} Distribution of edges across relation types.
\textbf{(d) Task definition.} SOTA discovery formulated as link prediction over the artifact graph.}
\label{fig:artifact-graph-vis}
\end{minipage}
\vspace{-4mm}
\end{figure}

\textbf{Thinking of HuggingFace as an artifact graph.}
We can conceptualize the HuggingFace ecosystem as a structured graph~\citep{Chen2025BenchmarkingRCA,Laufer2025AnatomyOAA}. As illustrated in Figure~\ref{fig:artifact-graph-vis}, models, datasets, papers, and codebases form nodes, while finetuning, reference, and evaluation relations form edges; evaluation metrics such as F1 scores serve as edge attributes. This graph view is enabled by three properties of HuggingFace: (\textbf{1}) a large and continuously growing artifact collection, (\textbf{2}) unified APIs that support programmatic access, and (\textbf{3}) rich metadata linking artifacts and evaluation results. Unlike academic papers, which often report results without directly linking them to executable resources, HuggingFace connects evaluation outcomes to concrete models and datasets, including many artifacts without formal publications. While prior work has mainly treated HuggingFace as a retrieval source~\citep{Silva2025ResearchKGA} or API hub~\citep{Shen2023TaskBenchBLA}, we study whether these structured artifact relationships can be converted to an engine for automatic discovery.

\textbf{Challenges for automatic discovery.}
Turning this artifact graph into an automatic discovery system presents two key challenges. (\textbf{1}) \textbf{Task ambiguity}: ``automatic discovery'' lacks a precise operational definition, making systematic evaluation difficult~\citep{Beel2025EvaluatingSAA}. (\textbf{2}) \textbf{Scalability}: the combinatorial number of possible model--dataset pairs makes exhaustive code-based verification prohibitively expensive~\citep{Urbanowicz2022STREAMLINEASA}.

\textbf{Linking artifacts as SOTA discovery.}
We address \textbf{task ambiguity} by instantiating automatic discovery as \textit{SOTA discovery}: identifying model--dataset pairs that achieve previously unseen evaluation scores. In the artifact graph, this becomes a concrete link prediction and ranking problem---identifying missing model--dataset links with high predicted edge attributes. This formulation provides a measurable objective for evaluating automatic discovery systems.

\textbf{Scalable framework for SOTA discovery.}
We address \textbf{scalability} with a two-stage \textit{rank-then-verify} framework. The ranking stage prioritizes promising model--dataset pairs using graph-based signals, substantially reducing the number of candidates that require expensive execution. The verification stage then synthesizes and executes evaluation code only for highly ranked pairs. This design enables efficient search while ensuring that discovered results are grounded in reproducible execution.

\textbf{Main contributions.}
Our work makes three contributions: (\textbf{1}) We construct \benchname, a benchmark for SOTA discovery in the HuggingFace ecosystem that includes prediction, ranking, and verification tasks; (\textbf{2}) We propose \modelname, a two-stage rank-then-verify framework and establish baseline results on \benchname; and (\textbf{3}) We demonstrate \modelname in an end-to-end Natural Language Inference (NLI) case study, showing its ability to discover new artifact relationships and generate research insights. Together, \benchname and \modelname provide a concrete benchmark and initial framework for studying automatic discovery over scientific artifacts.

\vspace{-1mm}
\section{Related Works}

\textbf{HuggingFace platform utilization.}
HuggingFace has increasingly become a natural platform for studying automatic discovery. {{Prior work has} largely relied on static analyses of its artifacts and relationships to characterize trends in machine learning development~\citep{Chen2025BenchmarkingRCA, Laufer2025AnatomyOAA}.} Beyond serving as a repository, HuggingFace has been conceptualized in multiple ways: as a knowledge graph~\citep{Silva2025ResearchKGA}, an API hub~\citep{Shen2023TaskBenchBLA}, a model card aggregator~\citep{Yang2024NavigatingDDA}, and even an evolutionary tree~\citep{Gao2023OnTOA}. Other studies have examined its community dynamics~\citep{Rahman2025HuggingGraphUTA, Castao2023AnalyzingTEA}. {In contrast, our work moves beyond static description and trend analysis and focuses on performance prediction and execution-based verification.}

\textbf{Large-scale prediction for accelerating discoveries.}
Accelerating scientific discovery has been a major focus in domains such as drug discovery~\citep{stokes2020deep,Serrano2024ArtificialIA,Vian2024IntegratingAIA,You2022ArtificialIIA}, materials science~\citep{xie2018crystal,butler2018machine}, and molecular design~\citep{segler2018planning}, {among others \citep{yu2025tinyscientist, cheng2025language}}.  In these settings, experimental verification is prohibitively costly and time-consuming. In contrast, our work focuses on a more tractable class of automatic discovery tasks by leveraging the intrinsic linking structure of HuggingFace artifacts.

\textbf{LLM-based coding {agents} for reproducible {experimentation}.}
Prior work has explored free-form discovery with generating executable code from research ideas~\citep{Lu2024TheASA, Jansen2024DISCOVERYWORLDAVA, Jansen2025CodeScientistESA}, though evaluation remains challenging given the open-ended nature of such tasks. Other efforts have focused on reproducing experiments within specific codebases~\citep{bogin2024super, starace2025paperbench,Kim2025FromRTA,Seo2025Paper2CodeACA,Siegel2024COREBenchFTA,Xiang2025SciReplicateBenchBLA, bragg2025astabench}, which is challenging due to the complexity of such codebases. In contrast, our tasks rely on reproducing a more concrete/grounded set of research artifacts.

\begin{table}[t]
\centering

\vspace{2pt}

\begin{threeparttable}
\small
\setlength{\tabcolsep}{4.0pt}

\begin{tabular}{@{}p{3.3cm}ccccccc@{}}
\toprule
\textbf{Resource}
& \textbf{Model}
& \textbf{Data}
& \textbf{Paper}
& \textbf{Code}
& \textbf{Eval. Score}
& \textbf{Model Lineage}
& \textbf{Execution} \\
\midrule

Linked PwC
& \checkmark
& \checkmark
& \checkmark
& \checkmark
& \checkmark
& \xmark
& \xmark \\

HuggingKG
& \checkmark
& \checkmark
& \checkmark
& \xmark
& \xmark
& \checkmark
& \xmark \\

HuggingGraph
& \checkmark
& \checkmark
& \xmark
& \xmark
& \xmark
& \checkmark
& \xmark \\

\midrule

\textbf{ArtifactBench}
& \checkmark
& \checkmark
& \checkmark
& \checkmark
& \checkmark
& \checkmark
& \checkmark \\

\bottomrule
\end{tabular}

\end{threeparttable}

\vspace{-2mm}
\caption{\textbf{Comparison with existing artifact-based benchmarks.}
Model, Data, Paper, and Code denote the included artifact types;
Eval. Score denotes model--dataset evaluation results with associated metrics and scores;
Model Lineage denotes model derivation relations such as fine-tuning;
and Execution denotes whether these benchmarks report reproduced evaluation results after execution.}
\label{tab:resource_comparison}
\vspace{-4mm}
\end{table}

\vspace{-1mm}
\section{Constructing an Artifact Graph from the HuggingFace Hub}
\vspace{-1mm}

We construct an artifact graph from the HuggingFace ecosystem to support
artifact discovery and execution-based verification.
Existing resources capture complementary aspects of the artifact ecosystem:
Linked PwC~\citep{farber2023linked} organizes models, datasets, papers, and benchmark
results, while HuggingKG~\citep{chen2025benchmarking} and HuggingGraph~\citep{rahman2025hugginggraph} capture HuggingFace artifacts and
their provenance relations.
However, as summarized in Table~\ref{tab:resource_comparison}, none jointly
provides model-dataset evaluation scores, model lineage, and executable
artifacts required for prediction followed by verification.
This motivates the artifact graph we construct below.

\textbf{Definition of artifact graphs.}
\label{artifact-graph-def}
We model the artifact ecosystem as a heterogeneous graph $\mathcal{G}=(\mathcal{V},\mathcal{E})$, where $\mathcal{V}=\mathcal{V}_m \cup \mathcal{V}_d \cup \mathcal{V}p \cup \mathcal{V}c$ contains four node types: models, datasets, papers, and codebases. Each node is associated with semantic attributes derived from its documentation, such as model cards, paper abstracts, and repository descriptions. The primary edge set, $\mathcal{E}_{\text{eval}} \subset \mathcal{V}_m \times \mathcal{V}_d$, represents evaluation relations: an edge $(m,d)$ indicates that model $m$ has been evaluated on dataset $d$ with observed score $f^*(m,d)$. These edges serve as both training supervision and prediction targets. We further include artifact--paper edges $\mathcal{E}_{\text{paper}}$, artifact--codebase edges $\mathcal{E}_{\text{code}}$, and model--model fine-tuning edges $\mathcal{E}_{\text{finetune}} \subset \mathcal{V}_m \times \mathcal{V}_m$. These auxiliary relations enrich the graph topology and provide additional structural context for message passing.

\textbf{Graph construction.}
We construct the artifact graph from popular models and datasets on the
HuggingFace Hub, with monthly downloads exceeding 1000, and enrich them with referenced arXiv papers and GitHub
repositories. The core of the graph consists of structured evaluation records
linking models to datasets with reported metrics and scores, while referenced
papers, codebases, and model lineage provide additional artifact relations.
After filtering isolated entries without any edges, the resulting graph
contains $14{,}053$ nodes and $51{,}337$ edges, with detailed statistics shown
in Figure~\ref{fig:artifact-graph-vis}. Artifact metadata is additionally
processed into leakage-free textual node representations for downstream
models. Full construction details and leakage analysis are provided in
Appendix~\S\ref{app:artifact-graph}.

\begin{figure*}[t]
    \centering
    \vspace{-2mm}
    \includegraphics[width=\linewidth]{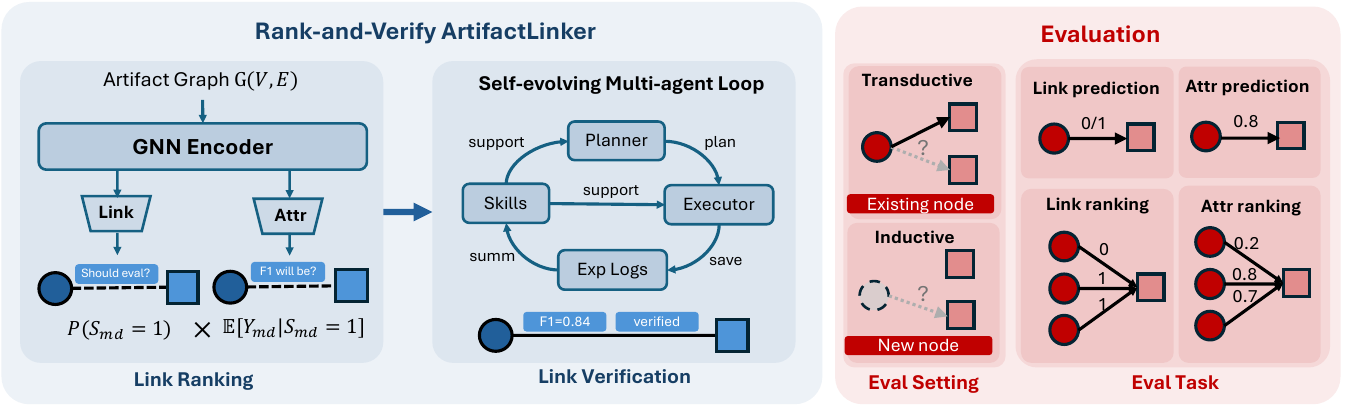}
    \caption{\textbf{Overview of \modelname and its evaluation framework.} \textbf{(Left) The two-stage rank-and-verify pipeline.} A GNN-based ranking model first estimates the ranking score for unobserved model--dataset pairs. The top-ranked candidates are then selected for execution in the verification stage. \textbf{(Right) Ranking evaluation tasks.} We evaluate the system under both transductive (nodes observed during training) and inductive (nodes unseen during training) settings. The evaluation spans four distinct tasks covering both link and attribute prediction/ranking.}
    \label{fig:method}
    \vspace{-2mm}
\end{figure*}

\vspace{-1mm}
\section{Linking Scientific Artifacts for Automatic SOTA Discovery}
\label{sec:link-artifact}
\vspace{-1mm}

We first formalize automatic SOTA discovery using the artifact-graph
formulation, and then introduce \modelname, a scalable two-stage
\textit{rank--and--verify} framework.

\vspace{-2mm}
\subsection{Definition of Automatic SOTA Discovery}
\label{sec:define-artifact}
\vspace{-2mm}

Building on the artifact graph
$\mathcal{G}=(\mathcal{V},\mathcal{E})$ defined in
Section~\S\ref{artifact-graph-def}, we treat each observed edge as a realized
evaluation record. A ground-truth oracle
$f^*: \mathcal{V}_m \times \mathcal{V}_d \rightarrow \mathbb{R}$
returns the benchmark score of a model--dataset pair. Automatic SOTA discovery
aims to identify a missing link $(m,d)\notin\mathcal{E}$ that, upon
verification, establishes a new state of the art by exceeding the current best
score on dataset $d$:
\begin{equation}
f^*(m,d) > f^*_{\max}(d)
= \max_{m' \mid (m',d)\in\mathcal{E}} f^*(m',d).
\end{equation}
In other words, the goal is to discover unobserved model--dataset pairs whose
verified performance is SOTA. The input to \modelname is the artifact graph
$\mathcal{G}$, and the output is a set of candidate $(m,d)$ pairs.

\textbf{Rank-and-verify framework.}
Evaluating all missing links through the code-execution oracle $f^*$ is
computationally prohibitive. We therefore adopt a two-stage pipeline: a
graph-based ranker first estimates $\hat{f}(m,d)$ for unobserved pairs
(\S\ref{link-ranker}), and a verifier then executes only the top-ranked
candidates to identify new SOTA results (\S\ref{link-verify}).

\vspace{-2mm}
\subsection{Link Ranking via Graph Modeling}
\label{link-ranker}

\textbf{The selection bias problem.}
The ranking stage estimates the expected performance $\hat{f}(m,d)$ of
unobserved model--dataset pairs without executing them. Let
$S_{md}\in\{0,1\}$ indicate whether a pair $(m,d)$ can be successfully
evaluated, yielding benchmark score $Y_{md}$. Training only on observed pairs
estimates the \emph{conditional} expectation
$\mathbb{E}[Y_{md}\mid S_{md}=1]$, implicitly assuming that the pair is viable.
However, ranking unobserved pairs requires the \emph{unconditional}
expectation $\mathbb{E}[Y_{md}]$, which must also account for model--dataset
incompatibility. Ignoring this distinction introduces selection bias and can
assign high predicted scores to pairs that ultimately fail to execute.

\textbf{Ranking formulation.}
We assume incompatible pairs yield zero utility, 
$\mathbb{E}[Y_{md}\mid S_{md}=0]=0$. By the law of total expectation,
\begin{equation}
\hat{f}(m,d)
= \mathbb{E}[Y_{md}]
= P(S_{md}=1)\cdot
  \mathbb{E}[Y_{md}\mid S_{md}=1].
\label{eq:score}
\end{equation}
This factorization decomposes ranking into two subproblems: a
\textit{link predictor} estimating the compatibility probability
$P(S_{md}=1)$, and an \textit{attribute predictor} estimating the conditional
performance $\mathbb{E}[Y_{md}\mid S_{md}=1]$.

\textbf{Ranking modeling.}
To instantiate this formulation, we first encode the artifact graph using an $L$-layer graph encoder:
\begin{equation}
\mathbf{h}_v^{(k+1)}
=
\mathrm{AGG}^{(k)}
\left(
\mathbf{h}_v^{(k)},
\left\{
\mathbf{h}_u^{(k)}, e_{uv}
: u\in\mathcal{N}(v)
\right\}
\right),
\end{equation}
yielding final node embeddings
$\mathbf{z}_v=\mathbf{h}_v^{(L)}$.
We then parameterize the two subproblems with two specialized heads over the
shared representations:
\begin{align}
P(S_{md}=1)
=
\phi_l(\mathbf{z}_m,\mathbf{z}_d),
\qquad
\mathbb{E}[Y_{md}\mid S_{md}=1]
=
\phi_a(\mathbf{z}_m,\mathbf{z}_d),
\label{eq:attr-head}
\end{align}
where $\phi_l$ is the link predictor, $\phi_a$ is the attribute predictor,
and $\epsilon$ captures observation noise. Both predictors jointly form to model the ranking function.

\textbf{Joint training.}
Let
$\hat{S}_{md}=\phi_l(\mathbf{z}_m,\mathbf{z}_d)$ denote predicted
compatibility and
$\hat{Y}_{md}=\phi_a(\mathbf{z}_m,\mathbf{z}_d)$ the predicted conditional
score. We train the link predictor using binary cross-entropy (BCE) over
observed positive pairs and randomly sampled negative pairs, while the
attribute predictor is trained using mean squared error (MSE) over
observed positive pairs:
\begin{align}
\mathcal{L}_{\text{link}}
=
\sum_{(m,d)\in\mathcal{E}^{+}\cup\mathcal{E}^{-}}
\mathrm{BCE}(\hat{S}_{md}, S_{md}), \qquad
\mathcal{L}_{\text{attr}}
=
\sum_{(m,d)\in\mathcal{E}^{+}}
\mathrm{MSE}(\hat{Y}_{md}, Y_{md}),
\end{align}
where $\mathcal{E}^{-}$ contains randomly sampled negative pairs.
Beyond realizing the factorization in Eq.~\eqref{eq:score}, the shared encoder
provides multi-task regularization. Although $\phi_a$ receives no direct
supervision on incompatible pairs, $\mathcal{L}_{\text{link}}$ shapes the
shared representations $\mathbf{z}$ with compatibility information, allowing
the attribute predictor to capture the viable--incompatible boundary beyond
what can be learned from $\mathcal{E}^{+}$ alone.

\textbf{Joint inference.}
At inference time, we rank each unobserved pair using
\begin{equation}
\hat{f}(m,d)
=
\phi_l(\mathbf{z}_m,\mathbf{z}_d)
\cdot
\phi_a(\mathbf{z}_m,\mathbf{z}_d).
\end{equation}
This multiplicative form naturally suppresses incompatible candidates:
as $\phi_l\rightarrow 0$, the final score approaches zero regardless of the
regressor output $\phi_a$. Among plausible pairs, $\phi_a$ then differentiates
candidates by their expected performance.
 
\vspace{-2mm}
\subsection{Link Verification with Self-Evolving Multi-Agent Framework}
\label{link-verify}

\textbf{The combination problem.}
Given the top-ranked candidate pairs $(m, d) \in \mathcal{C}$ collected from the ranking model $\hat{f}(m,d)$, the verification stage must automatically synthesize and execute code to verify the true benchmark score $Y_{md}$. However, standard coding agents treat each evaluation pair independently, causing repeated failures on shared artifact-specific constraints, such as atypical column schemas or required loading flags (e.g., \texttt{trust\_remote\_code=True}). We term this redundant rediscovery the \textit{combination problem}.

\textbf{Multi-agent self-evolution.}
We address this problem with a cross-instance memory loop built into a multi-agent framework. After each execution, an LLM reviewer distills runtime failures and successful workarounds into model-, dataset-, and task-specific memories. A \textit{planner agent} uses these accumulated memories to produce an error-aware evaluation plan, which an \textit{executor agent} implements and debugs through the CodeAct workflow \citep{wang2024executable}, with access to HuggingFace metadata APIs. This enables successful fixes to transfer across evaluation pairs that share previously encountered artifacts.

\vspace{-1mm}
\section{Experimental Settings}
\label{exp-settings}
We show ranking and verification task settings and baselines in this section. Full details about evaluation and baselines are available in Appendix~\S\ref{exp-details}.

\textbf{Ranking task settings.}
Our evaluation focuses on link and attribute ranking (Figure~\ref{fig:method}), while prediction metrics are also reported as a byproduct of the point-wise scoring function. We split the graph into train (80\%) and test (20\%) sets under two settings: (\textbf{1}) \textit{Transductive}, where edges are split while all nodes remain visible during training; and (\textbf{2}) \textit{Inductive}, where a subset of nodes is held out entirely to evaluate generalization to unseen artifacts. For attribute ranking, candidates are grouped by dataset and metric type.

\textbf{Ranking baselines.}
We compare against three groups of baselines: (\textbf{1}) \textbf{Heuristics}: Adamic-Adar, Katz, and Matrix Factorization; (\textbf{2}) \textbf{LLMs/Rankers}: Jina-v2-reranker\footnote{\url{https://huggingface.co/jinaai/jina-reranker-v2-base-multilingual}}, GPT-5.2~\citep{openai2025gpt52}, and Qwen3-8B~\citep{yang2025qwen3}; and (\textbf{3}) \textbf{GNNs}: GATv2Conv~\citep{velivckovic2017graph}, BUDDY-style~\citep{chamberlain2022graph}, NeoGNN-style~\citep{yun2021neo}, NCN~\citep{wang2023neural}, and NCNC~\citep{wang2023neural}. For LLMs and rerankers, we additionally evaluate \textit{+graph} variants that include the target nodes' 1-hop neighborhoods in the prompt. For a fair GNN comparison, all GNN methods use the same GATv2Conv encoder and differ only in their link-prediction decoders.

\begin{table*}[t]
\centering

\resizebox{\textwidth}{!}{%
\small
\setlength{\tabcolsep}{3.5pt}
\begin{tabular}{
  p{3cm}
  S[table-format=.3]
  S[table-format=.3]
  S[table-format=.3]
  S[table-format=.3]
  S[table-format=.3]
  S[table-format=.3]
  @{\hskip 8pt}
  S[table-format=.3]
  S[table-format=-.3]
  S[table-format=.3]
  S[table-format=.3]
  S[table-format=.3]
  S[table-format=.3]
}
\toprule
\multirow{3}{*}{\textbf{Method}}
& \multicolumn{6}{c}{\textbf{Transductive}}
& \multicolumn{6}{c}{\textbf{Inductive}} \\
\cmidrule(lr){2-7} \cmidrule(l){8-13}
& \multicolumn{2}{c}{Link Prediction}
& \multicolumn{4}{c@{\hskip 8pt}}{Link Ranking}
& \multicolumn{2}{c}{Link Prediction}
& \multicolumn{4}{c}{Link Ranking} \\
\cmidrule(lr){2-3} \cmidrule(lr){4-7} \cmidrule(lr){8-9} \cmidrule(l){10-13}
& {AP $\uparrow$} & {MCC $\uparrow$}
& {MRR $\uparrow$} & {Hit$_5$ $\uparrow$} & {R$_5$ $\uparrow$} & {N$_5$ $\uparrow$}
& {AP $\uparrow$} & {MCC $\uparrow$}
& {MRR $\uparrow$} & {Hit$_5$ $\uparrow$} & {R$_5$ $\uparrow$} & {N$_5$ $\uparrow$} \\
\midrule

\multicolumn{13}{l}{\textbf{Heuristic}}\\
Adamic-Adar
& .001 & .000 & .005 & .005 & .000 & .001
& .001 & .000 & .008 & .006 & .002 & .003 \\

Matrix-Factorization
& .015 & .000 & .052 & .066 & .017 & .028
& .002 & .000 & .001 & .000 & .000 & .000 \\

Katz
& .221 & {\cellcolor{bestgray}} .339
& .198 & .266 & .173 & .168
& .001 & -.001 & .004 & .002 & .002 & .002 \\

\midrule
\multicolumn{13}{l}{\textbf{Ranker-based}}\\
Jina-v2-reranker
& {--} & {--}
& .141 & .194 & .149 & .124
& {--} & {--}
& .144 & .190 & .137 & .121 \\

\rowcolor{graphrowcolor}
Jina-v2+graph
& {--} & {--}
& .246 & .301 & .211 & .201
& {--} & {--}
& .138 & .197 & .138 & .115 \\

\midrule
\multicolumn{13}{l}{\textbf{GNN-based w/ joint training}}\\
BUDDY-style
& .233 & .196
& .281 & .378 & .260 & .237
& .039 & .131
& .163 & .212 & .138 & .126 \\

NeoGNN-style
& .163 & .182
& .278 & .352 & .231 & .227
& .046 & .117
& {\cellcolor{bestgray}} .194 & .231 & .137 & .138 \\

NCN
& .314 & .207
& .297 & .389 & .256 & .248
& .057 & .147
& .165 & .218 & .140 & .132 \\

NCNC
& {\cellcolor{bestgray}} .352
& .241
& {\cellcolor{bestgray}} .309
& {\cellcolor{bestgray}} .398
& {\cellcolor{bestgray}} .266
& {\cellcolor{bestgray}} .259
& {\cellcolor{bestgray}} .063
& {\cellcolor{bestgray}} .162
& .173
& .229
& {\cellcolor{bestgray}} .159
& .145 \\

\rowcolor{graphrowcolor}
GATv2Conv
& .202 & .213
& .299 & .379 & .263 & .249
& .043 & .128
& .181
& {\cellcolor{bestgray}} .235
& .158
& {\cellcolor{bestgray}} .148 \\

\bottomrule
\end{tabular}%
}

\vspace{-2mm}
\caption{\textbf{Link prediction and ranking results.}
AP denotes PR-AUC. MCC is computed using a threshold selected on the validation set.
Hit$_5$, R$_5$, and N$_5$ denote Hits@5, Recall@5, and NDCG@5, respectively.
Jina-v2-reranker is built for ranking, so link prediction entries are left empty.
All GNN-based methods utilize joint training of links and attributes.
More details are provided in Section~\S\ref{exp-settings} and Appendix~\S\ref{exp-details}.}
\label{tab:link-prediction-ranking}

\vspace{3mm}

\resizebox{\textwidth}{!}{%
\small
\setlength{\tabcolsep}{3.5pt}
\begin{tabular}{
  p{3cm}
  S[table-format=.3]
  S[table-format=.3]
  S[table-format=.3]
  S[table-format=.3]
  S[table-format=.3]
  S[table-format=.3]
  @{\hskip 8pt}
  S[table-format=.3]
  S[table-format=.3]
  S[table-format=.3]
  S[table-format=.3]
  S[table-format=.3]
  S[table-format=.3]
}
\toprule
\multirow{3}{*}{\textbf{Method}}
& \multicolumn{6}{c}{\textbf{Transductive}}
& \multicolumn{6}{c}{\textbf{Inductive}} \\
\cmidrule(lr){2-7} \cmidrule(l){8-13}
& \multicolumn{2}{c}{Attr Prediction}
& \multicolumn{4}{c@{\hskip 8pt}}{Attr Ranking}
& \multicolumn{2}{c}{Attr Prediction}
& \multicolumn{4}{c}{Attr Ranking} \\
\cmidrule(lr){2-3} \cmidrule(lr){4-7} \cmidrule(lr){8-9} \cmidrule(l){10-13}
& {MAE $\downarrow$} & {RMSE $\downarrow$}
& {$\tau$ $\uparrow$} & {$\rho$ $\uparrow$} & {Hit$_1$ $\uparrow$} & {N$_1$ $\uparrow$}
& {MAE $\downarrow$} & {RMSE $\downarrow$}
& {$\tau$ $\uparrow$} & {$\rho$ $\uparrow$} & {Hit$_1$ $\uparrow$} & {N$_1$ $\uparrow$} \\
\midrule

\multicolumn{13}{l}{\textbf{Heuristic}}\\
Global mean
& .187 & .225
& .000 & .000 & .029 & .803
& .187 & .225
& .000 & .000 & .041 & .768 \\

Model mean
& .178 & .220
& .206 & .233 & .399 & .886
& .205 & .240
& .000 & .000 & .041 & .768 \\

Dataset mean
& .128 & .166
& .000 & .000 & .029 & .803
& .128 & .166
& .000 & .000 & .041 & .768 \\

\midrule
\multicolumn{13}{l}{\textbf{LLM-based}}\\
GPT-5.2
& .137 & .187
& .367 & .413 & .519 & .923
& .138 & .187
& .383 & .431 & .495 & .911 \\

\rowcolor{graphrowcolor}
GPT-5.2+graph
& .093 & .136
& .464 & .518
& {\cellcolor{bestgray}} .591
& .945
& .098 & .144
& {\cellcolor{bestgray}} .469
& {\cellcolor{bestgray}} .527
& {\cellcolor{bestgray}} .573
& {\cellcolor{bestgray}} .928 \\

Qwen3-8B
& .201 & .263
& .174 & .193 & .404 & .873
& .202 & .262
& .186 & .209 & .413 & .883 \\

\rowcolor{graphrowcolor}
Qwen3-8B+graph
& .131 & .181
& .207 & .246 & .452 & .898
& .121 & .173
& .205 & .232 & .395 & .876 \\

\midrule
\multicolumn{13}{l}{\textbf{GNN-based w/ joint training}}\\
BUDDY-style
& .064 & .101
& .470 & .531 & .573
& {\cellcolor{bestgray}} .957
& .098 & .141
& .427 & .485 & .453 & .908 \\

NeoGNN-style
& .064 & .100
& .483 & .549 & .542 & .946
& .095
& {\cellcolor{bestgray}} .135
& .362 & .423 & .418 & .911 \\

NCN
& .064 & .101
& .447 & .508 & .520 & .950
& .097 & .138
& .448 & .515 & .498 & .919 \\

NCNC
& .064 & .100
& .433 & .496 & .520 & .938
& .097 & .138
& .402 & .466 & .449 & .917 \\

\rowcolor{graphrowcolor}
GATv2Conv
& {\cellcolor{bestgray}} .063
& {\cellcolor{bestgray}} .099
& {\cellcolor{bestgray}} .508
& {\cellcolor{bestgray}} .572
& .560
& .955
& {\cellcolor{bestgray}} .094
& .136
& .387
& .450
& .458
& .915 \\

\bottomrule
\end{tabular}%
}

\vspace{-2mm}
\caption{\textbf{Attribute prediction and ranking results.}
$\tau$, $\rho$, Hit$_1$, and N$_1$ denote Kendall's Tau, Spearman's Rho,
Hit@1, and NDCG@1, respectively.
All GNN-based methods utilize joint training of links and attributes.
More details are provided in Section~\S\ref{exp-settings} and Appendix~\S\ref{exp-details}.}
\label{tab:attr-prediction-ranking}

\vspace{-2mm}
\end{table*}

\textbf{Verification task settings.}
We construct a subset of the benchmark of 263 model--dataset pairs from the test set for automated evaluation reproduction. We filter pairs using three criteria: (\textbf{1}) prioritize popular models and datasets by download count; (\textbf{2}) exclude artifacts that exceed practical execution constraints; and (\textbf{3}) restrict evaluation to bounded metrics---\texttt{accuracy}, \texttt{F1}, \texttt{BLEU}, \texttt{chrF}, and \texttt{rouge}---to ensure consistent evaluation. For each pair, the agent must reproduce the officially reported metric through fully automated code execution.

\textbf{Verification baselines.}
We evaluate the contribution of each agent component using four baselines:
(\textbf{1}) \textbf{Agent-free}: generation without iterative execution feedback;
(\textbf{2}) \textbf{ReAct agent}: iterative reasoning without HuggingFace API tools;
(\textbf{3}) \textbf{Tool-use agent}: a single agent responsible for both planning and coding; and
(\textbf{4}) \textbf{Tool-use multi-agent}: a planner--executor framework without self-evolving memory.

\begin{figure}[t]
    \centering
    \begin{minipage}[t][10.1cm][t]{0.31\linewidth}
        \centering
        \includegraphics[width=\linewidth]{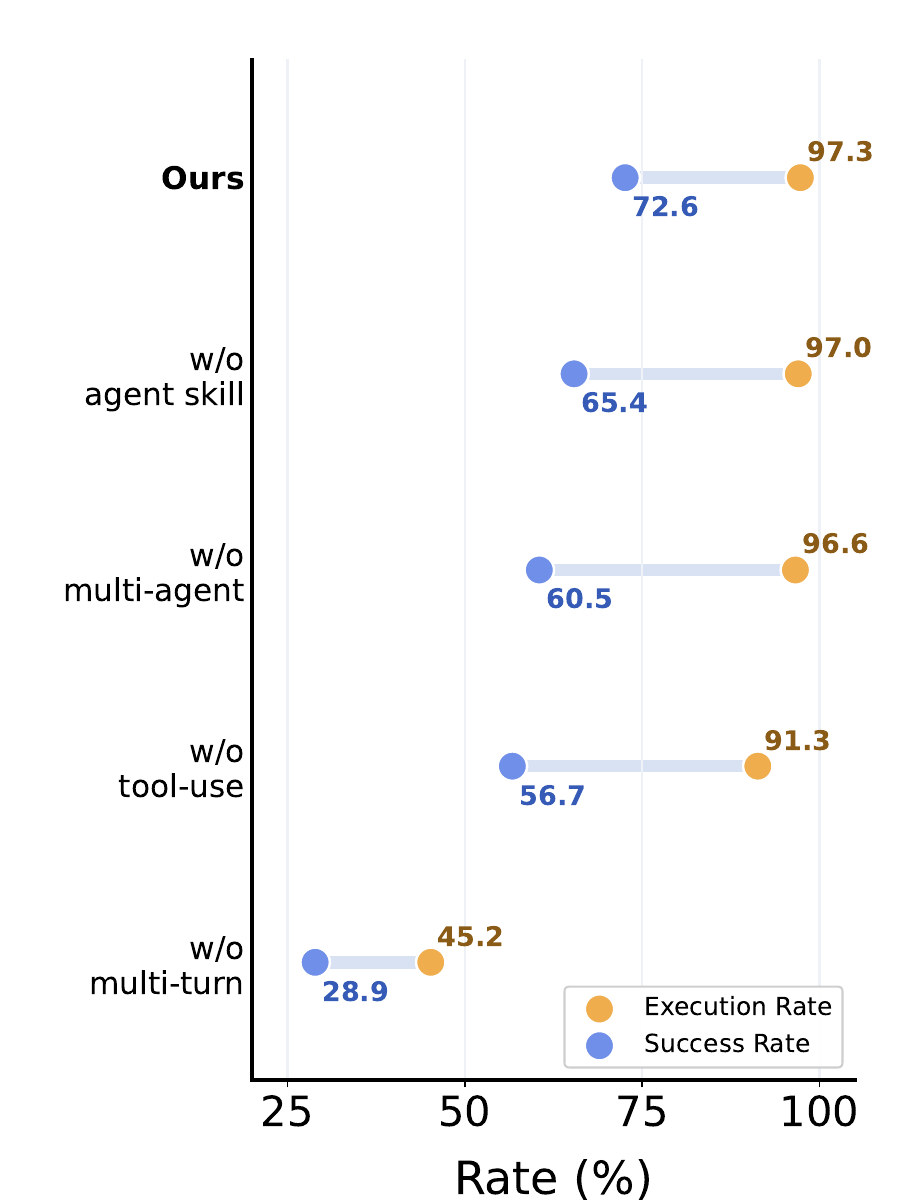}
        \caption{\textbf{Agentic verification performance under component ablations.} We ablate each component in our method. Execution rate measures whether the generated code runs successfully, and success rate measures whether the result is no worse than 80\% of the ground-truth performance.}
        \label{fig:artifact-coder}
    \end{minipage}\hfill
    \begin{minipage}[t][10.1cm][t]{0.31\linewidth}
        \centering
        \includegraphics[width=\linewidth]{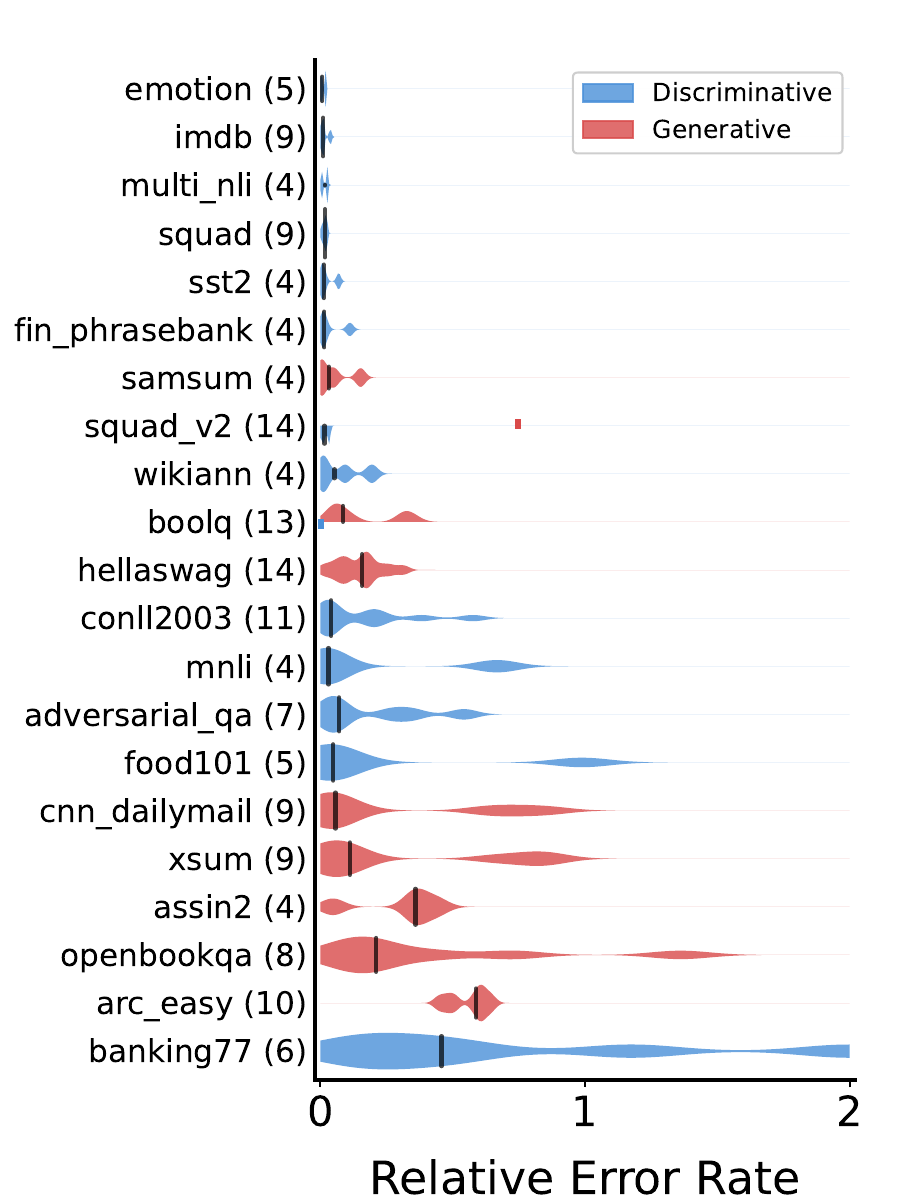}
        \caption{\textbf{Error distribution of reproduced verification results.} We show the error distribution across datasets in our reproduced evaluation. The number after each dataset name denotes the number of evaluated models, and discriminative and generative models are shown separately.}
        \label{fig:artifact-coder-analysis}
    \end{minipage}\hfill
    \begin{minipage}[t][10.1cm][t]{0.31\linewidth}
        \centering
        \includegraphics[width=\linewidth]{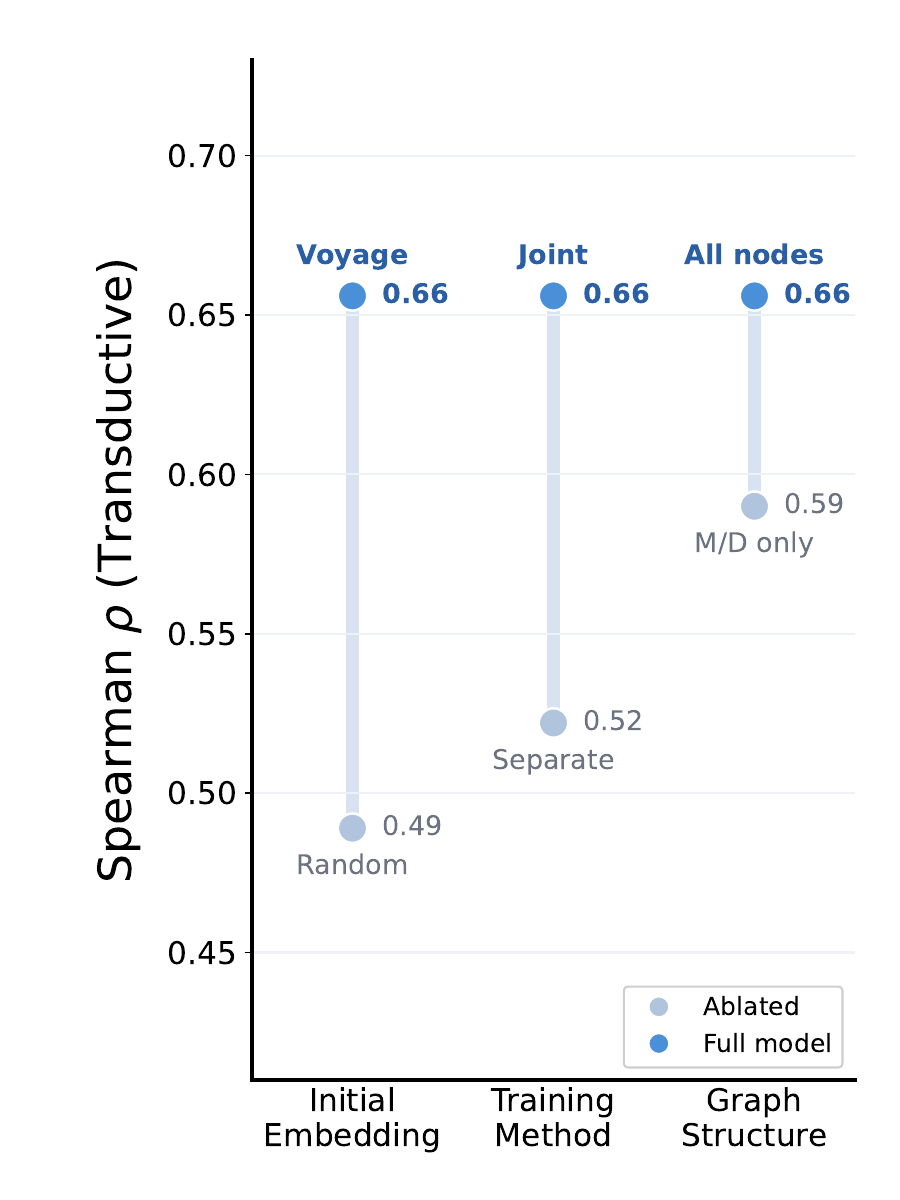}
        \caption{\textbf{Ablation study of the GNN-based ranker.} We study three factors: (1) encoder embedding initialization (Voyage vs.\ random), (2) training strategy (dual-head joint training vs.\ independent training), and (3) graph structure (model--dataset nodes only vs.\ the full graph).}
        \label{fig:predictor-ablation}
    \end{minipage}
\end{figure}

\section{Experimental Results}

\textbf{GNN-based ranking models match or outperform LLM-based methods.}
Tables~\ref{tab:link-prediction-ranking} and \ref{tab:attr-prediction-ranking} show that specialized GNNs are highly competitive when graph structure is available. For transductive link ranking, NCNC achieves the best MRR of 0.309, outperforming the strongest reranker baseline, Jina-v2+graph (0.246). GNNs are also particularly strong for attribute prediction: GATv2Conv achieves an MAE of 0.063, compared with 0.093 for GPT-5.2+graph, while also obtaining the best Kendall's $\tau$ and Spearman's $\rho$. Overall, these results demonstrate the value of structural information in the artifact graph.

\textbf{GNNs retain strong performance under inductive evaluation.}
Although performance generally decreases from transductive to inductive settings, GNNs remain competitive with LLM-based methods. For inductive attribute prediction, GATv2Conv achieves the best MAE of 0.094, outperforming GPT-5.2+graph at 0.098, while NeoGNN achieves the best RMSE of 0.135. For attribute ranking, NCN reaches a Kendall's $\tau$ of 0.448 and Spearman's $\rho$ of 0.515, close to GPT-5.2+graph at 0.469 and 0.527. These results suggest that the learned structural representations generalize to unseen artifact combinations rather than memorizing.

\textbf{ArtifactCoder reliably reproduces evaluations for widely used artifacts.}
Beyond prediction on the artifact graph, Figure~\ref{fig:artifact-coder} shows that ArtifactCoder achieves a 72.6\% success rate across 263 evaluations. Multi-turn interaction is particularly important: removing it reduces success to 28.9\%, while removing tool use lowers performance to 56.7\%. Together, self-evolving memory, multi-agent coordination, and tool use substantially improve the reliability of automatic evaluation reproduction.

\textbf{Link verification in the wild provides a challenging agent benchmark.}
Figure~\ref{fig:artifact-coder-analysis} reveals substantial variation across artifacts. ArtifactCoder achieves near-zero error on common datasets such as SQuAD and IMDb, but struggles with more complex or less frequently used artifacts. ARC-Easy exhibits consistently high relative error across models, while Banking77 frequently exceeds 0.5 and can reach 2.0. These results highlight in-the-wild artifact verification as a challenging setting for evaluating agent reasoning, debugging, and tool-use capabilities, particularly on complex and niche artifacts. As new artifacts continue to emerge, the benchmark can also naturally expand over time.

\vspace{-1mm}
\section{Discussions}

\begin{figure}[t]
    \centering
    \begin{minipage}[t][10.1cm][t]{0.31\linewidth}
        \centering
        \includegraphics[width=\linewidth]{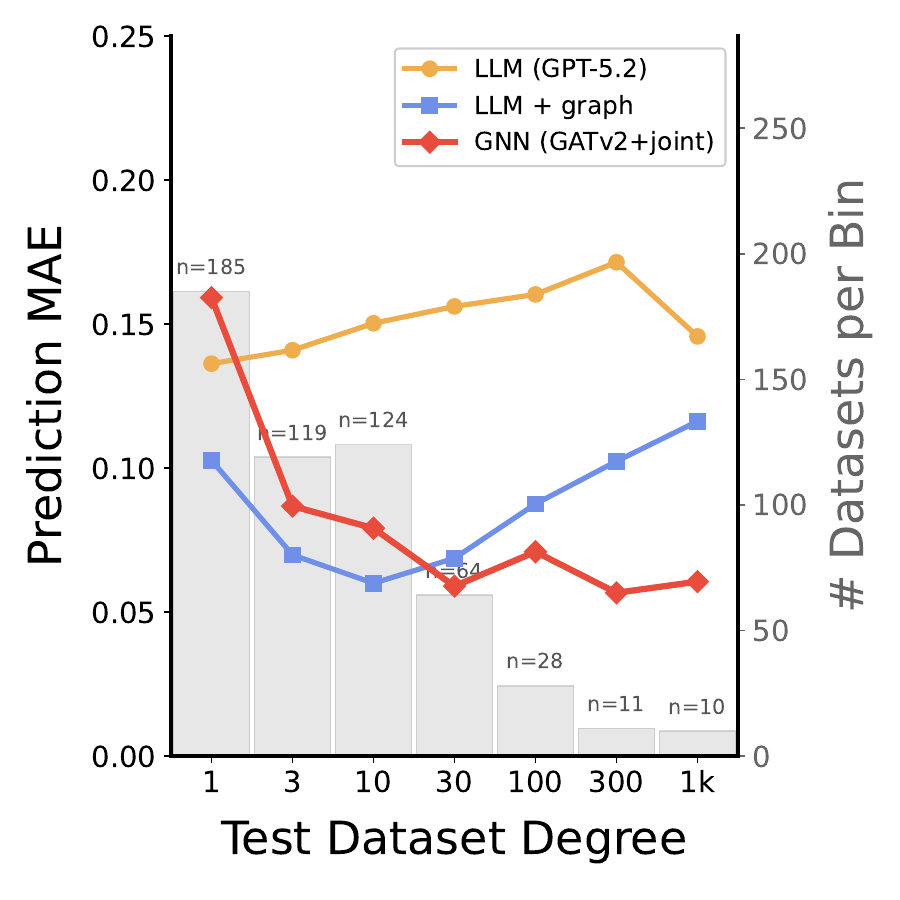}
        \caption{\textbf{Degree analysis of attribution prediction results.} We ablate on LLMs, LLMs with 1-hop neighborhood context, and GNN-based methods. We split the test set based on the node degrees of the datasets. Gray bars indicate the degree distribution of dataset nodes.}
        \label{fig:degree-analysis}
    \end{minipage}\hfill
    \begin{minipage}[t][10.1cm][t]{0.31\linewidth}
        \centering
        \includegraphics[width=\linewidth]{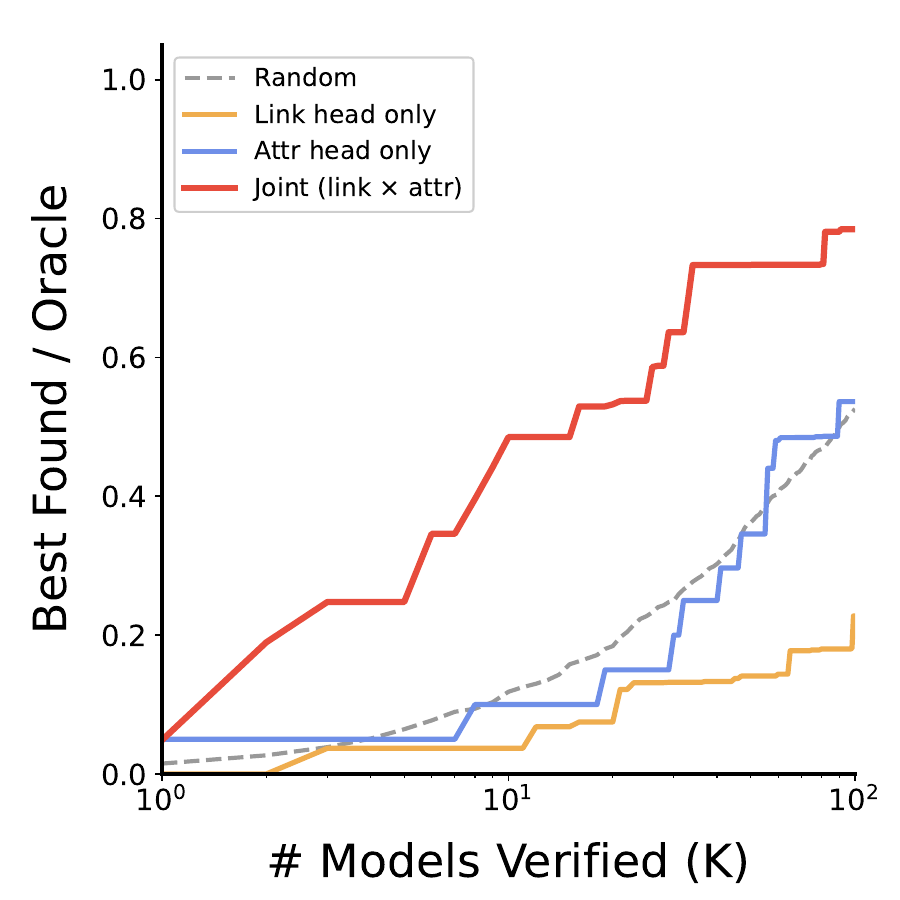}
        \caption{\textbf{Verification cost reduction via joint ranking.} We rank candidate models by different scoring functions and verify them in rank order. The y-axis shows the best performance found (normalized by the oracle) after verifying the top-K models (x-axis).}
        \label{fig:ranking-cost-curve}
    \end{minipage}\hfill
    \begin{minipage}[t][10.1cm][t]{0.31\linewidth}
        \centering
        \includegraphics[width=\linewidth]{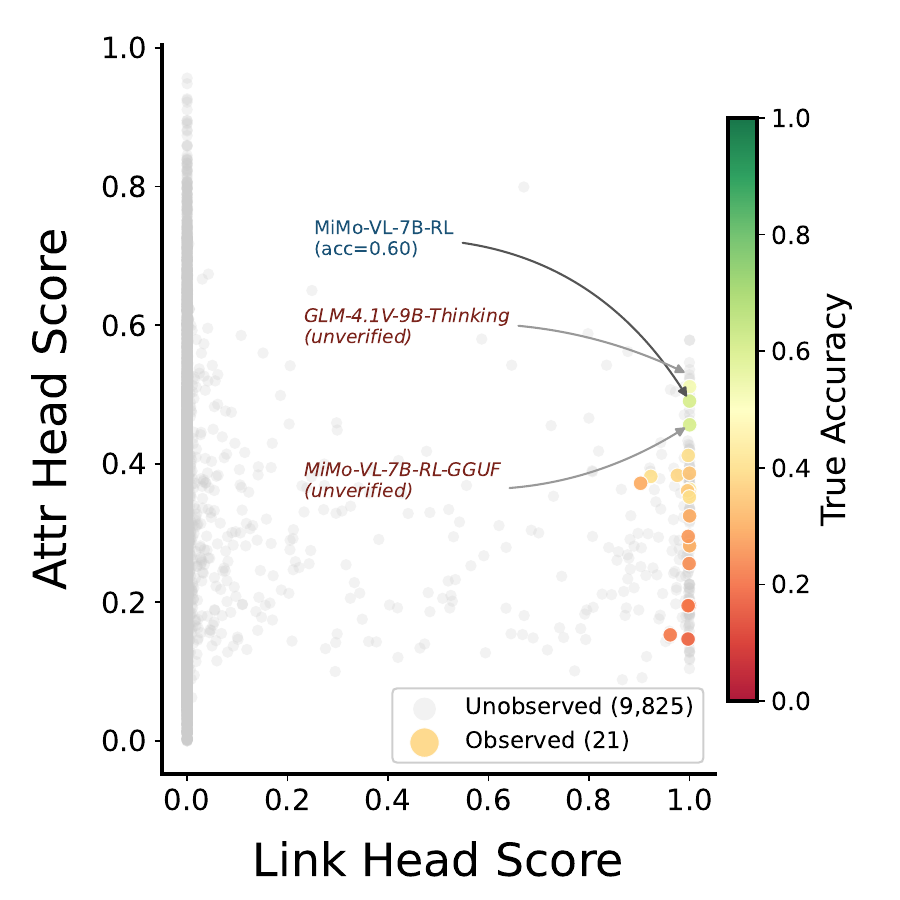}
        \caption{\textbf{Case study of model ranking.} We show an example ranking of all models on MathVision. Each dot represents a model, positioned by its link head score (x-axis) and attribute head score (y-axis). Gray dots denote models without existing reported evaluations.}
        \label{fig:ranking-case-study}
    \end{minipage}
\vspace{-17mm}
\end{figure}

\textbf{Q1: Which factors support effective ranking in the artifact graph?}
The effectiveness of our GNN framework comes from both informative graph representations and joint training. As shown in Figure~\ref{fig:predictor-ablation}, stronger initial semantic embeddings provide the largest gain, increasing correlation from 0.49 to 0.66. Adding richer graph structure, including paper and codebase nodes, further improves performance from 0.59 to 0.66. To avoid data leakage, all metadata features are derived from LLM-summarized model cards with ground-truth labels removed. More detailed leakage analysis is in Appendix~\S\ref{app:artifact-graph}. Finally, shared-encoder joint training improves correlation from 0.52 to 0.66. Together, these results show that effective ranking depends on combining semantic node features, graph topology, and joint optimization.

\textbf{Q2: Why can GNN-based methods outperform LLM-based methods in attribute prediction?}
Across all evaluations, the joint GATv2 model achieves an MAE of 0.062, outperforming GPT-5.2 alone (0.137) and GPT-5.2 with 1-hop graph context (0.093). Figure~\ref{fig:degree-analysis} shows that this advantage grows with node degree: GNN MAE decreases monotonically from 0.087 at degree 3--10 to 0.057 at degree 300--1,000 as richer neighborhoods become available. In contrast, LLM+graph degrades to 0.102--0.116 at high degree, where serializing large neighborhoods requires truncation due to context-window limits. At cold start (degree 1--3, ($\sim$34\%) of datasets), however, both LLM variants outperform GNNs (MAE 0.103--0.136 vs.\ 0.159), because textual priors remain useful when graph evidence is sparse. This suggests a natural hybrid strategy: use LLM+graph for cold-start datasets and GNNs for well-connected artifacts.

\textbf{Q3: How well does the ranking model reduce verification costs?}
The ranking model $\hat{f}(m,d)$ is designed to prioritize promising SOTA candidates and reduce verification cost. As shown in Figure~\ref{fig:ranking-cost-curve}, ranking by $\hat{f}(m,d)$ recovers 50\% of existing SOTA results with only 10 verifications on average, compared with more than 60 for the biased attribution baseline. Figure~\ref{fig:ranking-case-study} further shows a case study on MathVision, where the model ranks highly both variants of known strong models (e.g., MiMo-VL-7B-RL-GGUF\footnote{\url{https://huggingface.co/unsloth/MiMo-VL-7B-RL-GGUF}}) and distinct model families (e.g., GLM-4.1V\footnote{\url{https://huggingface.co/zai-org/GLM-4.1V-9B-Base}}), demonstrating its ability to generalize beyond model variants.

\begin{figure*}[t]
    \begin{minipage}[t]{0.31\linewidth}
        \centering
        \includegraphics[width=\linewidth]{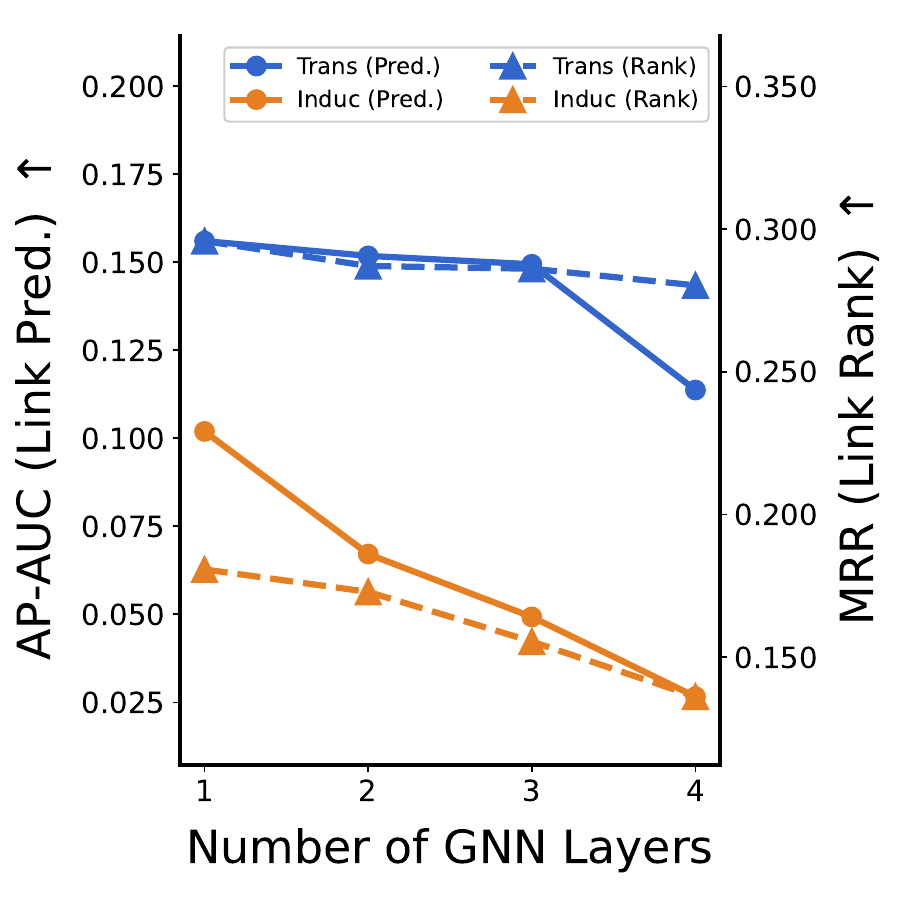}
        \captionof{figure}{\textbf{Ablation study on GNN layer numbers (link ranking and prediction).} GATv2 as the backbone model. Both AP-AUC and MRR metrics are the higher the better.}
        \label{fig:ablation-layer-link}
    \end{minipage}\hfill
    \begin{minipage}[t]{0.31\linewidth}
        \centering
        \includegraphics[width=\linewidth]{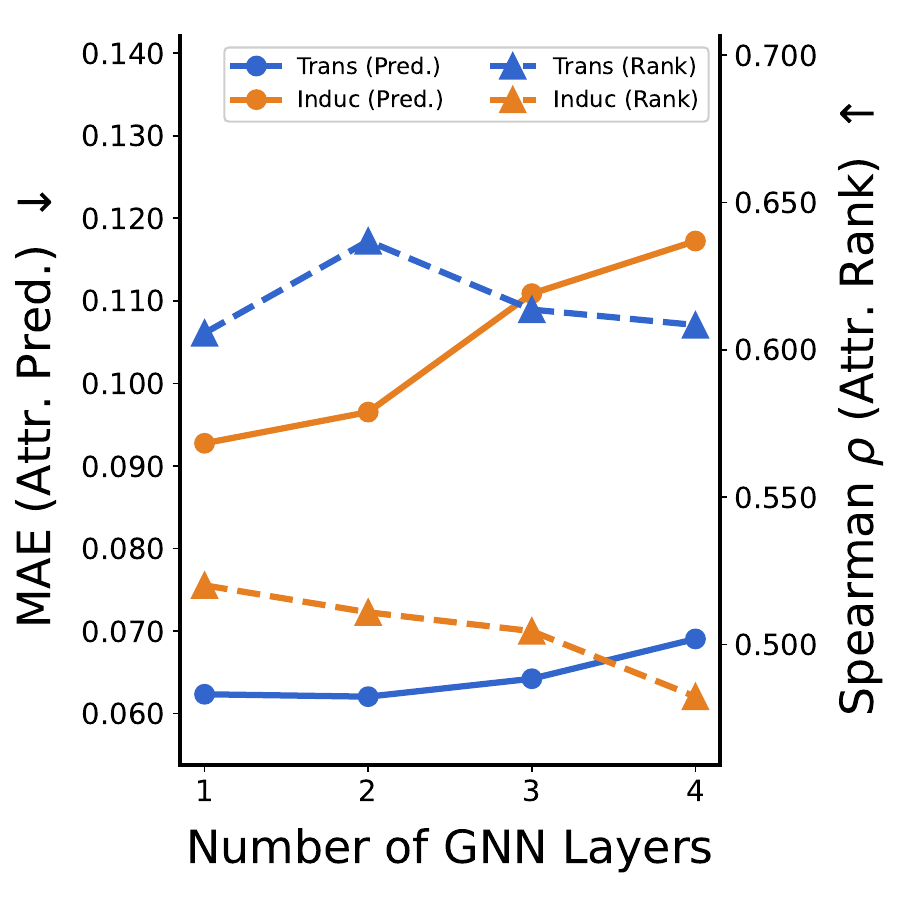}
        \captionof{figure}{\textbf{Ablation study on GNN layer numbers (attribute ranking and prediction).} GATv2 as the backbone model. MAE is the lower the better while Spearman is the higher the better.}
        \label{fig:ablation-layer-attr}
    \end{minipage}\hfill
    \begin{minipage}[t]{0.31\linewidth}
        \centering
        \includegraphics[width=\linewidth]{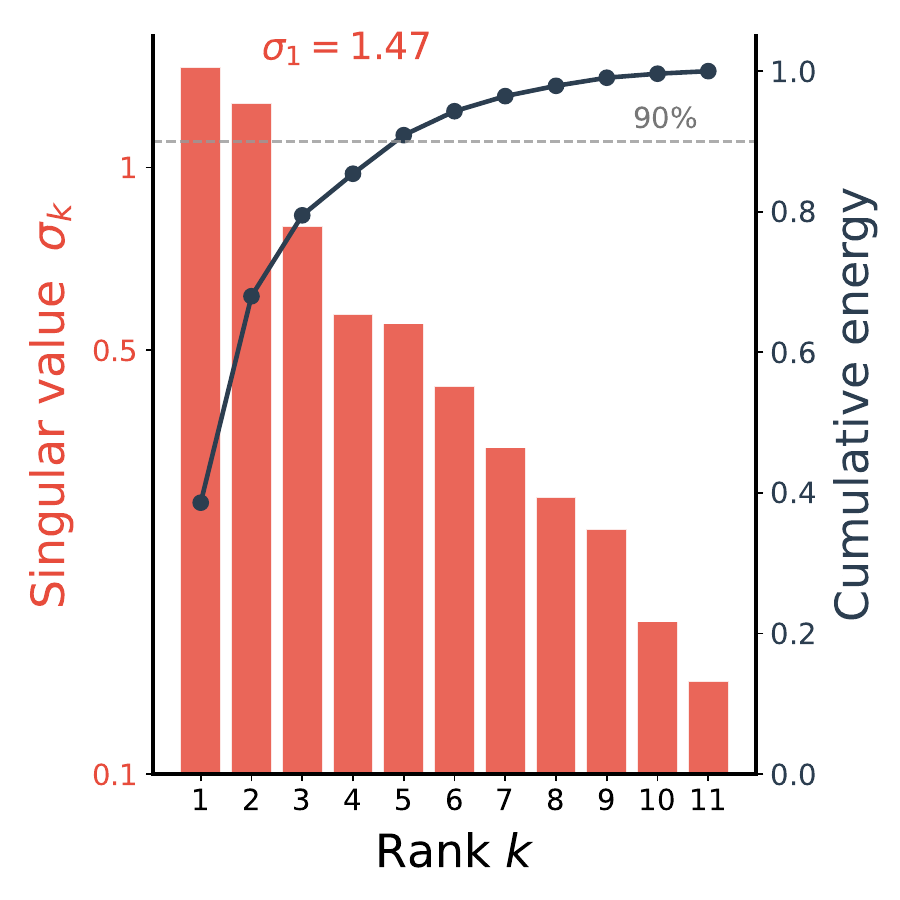}
        \captionof{figure}{\textbf{Rank analysis of NLI accuracy matrix.} SVD applied to the double-centered 12 × 45 dataset × model accuracy matrix. Rank-5 recovers ~90\%+ of the residual variance.}
        \label{fig:ranking-analysis}
    \end{minipage}
    \vspace{-2mm}
\end{figure*}

\vspace{-1mm}
\section{Ablation Studies}

\textbf{GNN decoders.}
Besides the bilinear decoder used in GATv2, we evaluate specialized link-prediction decoders, including NCN, NCNC, NeoGNN, and BUDDY, while keeping the GATv2 encoder fixed. These decoders exploit additional neighborhood structure and generally improve link prediction and ranking, with NCNC achieving the strongest transductive link-ranking performance. Their benefits are less consistent on attribute prediction and ranking, where the same node representations are shared across variants. This suggests that explicit structural priors are particularly effective for modeling link compatibility, while fine-grained attribute estimation depends more on the learned node representations.

\textbf{GNN layer number.}
In Figures~\ref{fig:ablation-layer-link} and \ref{fig:ablation-layer-attr}, we vary the number of GNN layers from 1 to 4. Both link- and attribute-level performance peaks at one or two layers and degrades thereafter, particularly on the inductive split: AP drops by half from (L=1) to (L=4), while attribute MAE increases by over 25\%. This is consistent with over-smoothing around high-degree dataset hubs. Nevertheless, graph aggregation remains essential: removing it entirely (matrix factorization with text embeddings) reduces transductive link MRR from 0.30 to 0.01 and attribute $\rho$ from 0.61 to 0.41. Thus, a shallow non-linear GNN is sufficient to capture the local collaborative-filtering signal without excessive over-smoothing.

\begin{figure}[t]
    \centering
    \includegraphics[width=\linewidth]{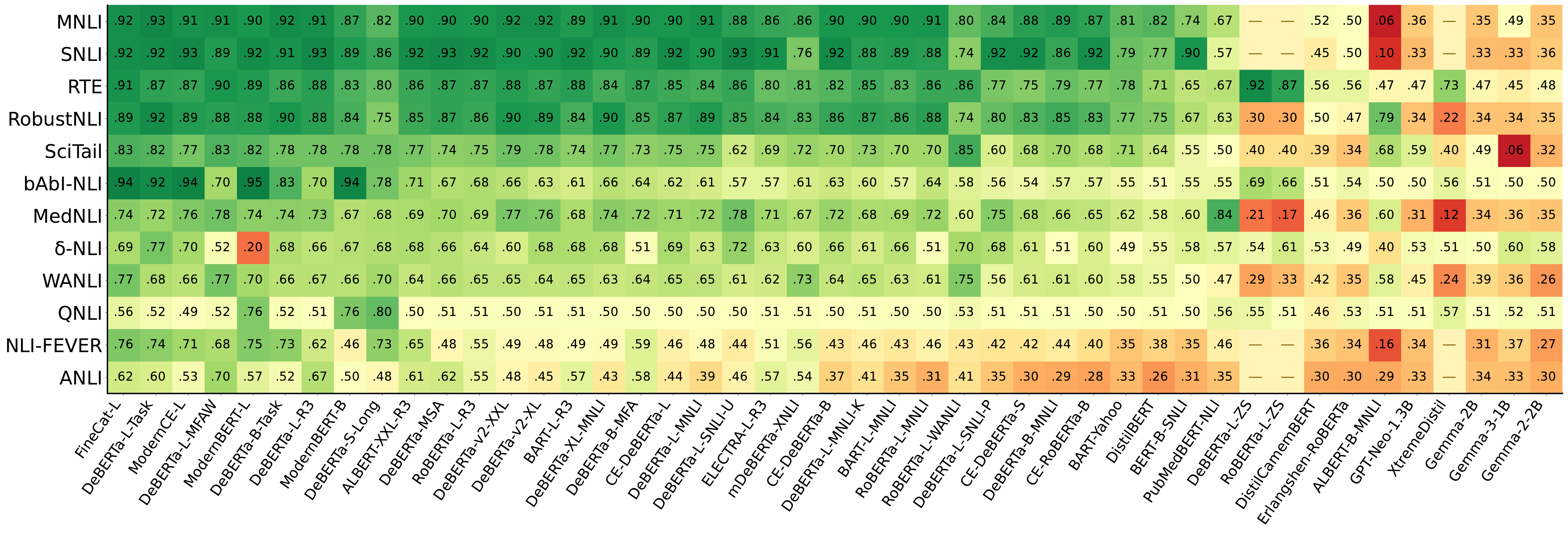}
    \vspace{-6mm}
    \caption{\textbf{Verified accuracy matrix for NLI tasks.} We show the accuracy verification results conducted by the \modelname with 45 models and 12 NLI datasets. We use ST\_SE split for RobustNLI. Cells filled with "--" are because these models are two-way pretrained models, while the evaluated datasets are 3-way NLI tasks. Therefore, these models are skipped for typical datasets. Models and datasets details are in Appendix~\S\ref{appendix:case-study-detail}.}
    \label{fig:nli-results-heatmap}
\end{figure}

\vspace{-2mm}
\section{End-to-end Case Study}
\vspace{-2mm}
\label{sec:case-study}

We demonstrate \modelname{} on Natural Language Inference (NLI), evaluating 45 candidate models across 12 representative NLI datasets \citep{mnli, snli, nie2020adversarial, xnli, rte, wang2018glue} (Figure~\ref{fig:nli-results-heatmap}). We choose NLI because its rich collection of models, datasets, and published results provides a well-grounded setting for end-to-end evaluation. The case study illustrates the types of discoveries that \modelname{} can extract from an artifact graph.

\textbf{Q1: Can \modelname{} discover unobserved SOTA results in the real world?}
We identify \texttt{sileod/deberta-v3-large-tasksource-nli},\footnote{\url{https://huggingface.co/sileod/deberta-v3-large-tasksource-nli}} a multi-task model whose card reports results on WNLI and MNLI but not SNLI. Our verification obtains an SNLI test accuracy of 0.9212---to our knowledge, the first reported score for this model and within 1,pp of the leaderboard SOTA (0.931).\footnote{\url{https://nlp.stanford.edu/projects/snli/}} On the less canonical \texttt{pietrolesci/robust\_nli} (ST\_SE),\footnote{\url{https://huggingface.co/datasets/pietrolesci/robust_nli}} the same model reaches 0.920, improving over the previously reported best by 23pp. These results show that \modelname{} can uncover strong, previously unreported evaluations.

\textbf{Q2: Does the model--dataset evaluation matrix exhibit low-rank hidden structure?}
Applying SVD to the double-centered $12\times45$ accuracy matrix shows that three components explain 80\% of the residual variance and five explain 91\% (Figure~\ref{fig:ranking-analysis}), indicating an effective interaction rank of only $3$--$5$. This helps explain why a shallow GNN is sufficient: much of the model--dataset interaction can be captured in a low-dimensional space. Consistent with this structure, \modelname's GNN-based factorization achieves 0.062 MAE over 5{,}839 transductive edges. The completed evaluation matrix also reveals non-monotonic model capabilities: newer models, including the Gemma family, often underperform older DeBERTa-based models on NLI tasks that the latter solve near-perfectly. Although similar phenomena have been studied in prior NLI work \citep{naik2018stress,talman2019testing,bhargava2021generalization,delbari2025beyond}, \modelname{} recovers these patterns directly from large-scale evaluation traces, illustrating its ability to support automated empirical analysis.

\vspace{-1mm}
\section{Conclusion}
\vspace{-1mm}
In this paper, we introduce \benchname, a challenging benchmark for SOTA discovery grounded in the HuggingFace ecosystem. We also present \modelname, a two-stage rank-and-verify framework, together with an end-to-end Natural Language Inference case study demonstrating its practical potential in realistic discovery settings. Our results establish \benchname as a meaningful testbed for automated scientific discovery and highlight the promise of rank-and-verify approaches for efficiently exploring large discovery spaces, while providing a foundation for future tasks and research in this area.

\bibliography{custom}

@inproceedings{naik2018stress,
  title={Stress test evaluation for natural language inference},
  author={Naik, Aakanksha and Ravichander, Abhilasha and Sadeh, Norman and Rose, Carolyn and Neubig, Graham},
  booktitle={Proceedings of the 27th International Conference on Computational Linguistics},
  pages={2340--2353},
  year={2018}
}

@inproceedings{rahman2025hugginggraph,
  title={Hugginggraph: Understanding the supply chain of llm ecosystem},
  author={Rahman, Mohammad Shahedur and Gao, Peng and Ji, Yuede},
  booktitle={Proceedings of the 34th ACM International Conference on Information and Knowledge Management},
  pages={5997--6005},
  year={2025}
}

@inproceedings{chen2025benchmarking,
  title={Benchmarking recommendation, classification, and tracing based on hugging face knowledge graph},
  author={Chen, Qiaosheng and Huang, Kaijia and Zhou, Xiao and Luo, Weiqing and Cui, Yuanning and Cheng, Gong},
  booktitle={Proceedings of the 48th International ACM SIGIR Conference on Research and Development in Information Retrieval},
  pages={3433--3443},
  year={2025}
}

@article{farber2023linked,
  title={Linked papers with code: the latest in machine learning as an RDF knowledge graph},
  author={F{\"a}rber, Michael and Lamprecht, David},
  journal={arXiv preprint arXiv:2310.20475},
  year={2023}
}

@article{yang2025qwen3,
  title={Qwen3 technical report},
  author={Yang, An and Li, Anfeng and Yang, Baosong and Zhang, Beichen and Hui, Binyuan and Zheng, Bo and Yu, Bowen and Gao, Chang and Huang, Chengen and Lv, Chenxu and others},
  journal={arXiv preprint arXiv:2505.09388},
  year={2025}
}

@article{chamberlain2022graph,
  title={Graph neural networks for link prediction with subgraph sketching},
  author={Chamberlain, Benjamin Paul and Shirobokov, Sergey and Rossi, Emanuele and Frasca, Fabrizio and Markovich, Thomas and Hammerla, Nils and Bronstein, Michael M and Hansmire, Max},
  journal={arXiv preprint arXiv:2209.15486},
  year={2022}
}

@article{wang2023neural,
  title={Neural common neighbor with completion for link prediction},
  author={Wang, Xiyuan and Yang, Haotong and Zhang, Muhan},
  journal={arXiv preprint arXiv:2302.00890},
  year={2023}
}

@article{velivckovic2017graph,
  title={Graph attention networks},
  author={Veli{\v{c}}kovi{\'c}, Petar and Cucurull, Guillem and Casanova, Arantxa and Romero, Adriana and Lio, Pietro and Bengio, Yoshua},
  journal={arXiv preprint arXiv:1710.10903},
  year={2017}
}

@inproceedings{talman2019testing,
  title={Testing the generalization power of neural network models across NLI benchmarks},
  author={Talman, Aarne and Chatzikyriakidis, Stergios},
  booktitle={Proceedings of the 2019 ACL workshop BlackboxNLP: Analyzing and interpreting neural networks for NLP},
  pages={85--94},
  year={2019}
}

@inproceedings{yu2025tinyscientist,
  title={Tinyscientist: An interactive, extensible, and controllable framework for building research agents},
  author={Yu, Haofei and Xuan, Keyang and Li, Fenghai and Zhu, Kunlun and Lei, Zijie and Zhang, Jiaxun and Qi, Ziheng and Richardson, Kyle and You, Jiaxuan},
  booktitle={Proceedings of the 2025 Conference on Empirical Methods in Natural Language Processing: System Demonstrations},
  pages={558--590},
  year={2025}
}

@article{kingma2014adam,
  title={Adam: A method for stochastic optimization},
  author={Kingma, Diederik P and Ba, Jimmy},
  journal={arXiv preprint arXiv:1412.6980},
  year={2014}
}

@inproceedings{bhargava2021generalization,
  title={Generalization in NLI: Ways (not) to go beyond simple heuristics},
  author={Bhargava, Prajjwal and Drozd, Aleksandr and Rogers, Anna},
  booktitle={Proceedings of the Second Workshop on Insights from Negative Results in NLP},
  pages={125--135},
  year={2021}
}

@inproceedings{delbari2025beyond,
  title={Beyond accuracy: Revisiting out-of-distribution generalization in NLI models},
  author={Delbari, Zahra and Pilehvar, Mohammad Taher},
  booktitle={Proceedings of the 29th Conference on Computational Natural Language Learning},
  pages={557--570},
  year={2025}
}

@inproceedings{wang2018glue,
  title={GLUE: A multi-task benchmark and analysis platform for natural language understanding},
  author={Wang, Alex and Singh, Amanpreet and Michael, Julian and Hill, Felix and Levy, Omer and Bowman, Samuel},
  booktitle={Proceedings of the 2018 EMNLP workshop BlackboxNLP: Analyzing and interpreting neural networks for NLP},
  pages={353--355},
  year={2018}
}

@article{rte,
  title={The Fifth PASCAL Recognizing Textual Entailment Challenge.},
  author={Bentivogli, Luisa and Clark, Peter and Dagan, Ido and Giampiccolo, Danilo},
  journal={TAC},
  volume={7},
  number={8},
  pages={1},
  year={2009}
}

@inproceedings{xnli,
  title={XNLI: Evaluating cross-lingual sentence representations},
  author={Conneau, Alexis and Rinott, Ruty and Lample, Guillaume and Williams, Adina and Bowman, Samuel and Schwenk, Holger and Stoyanov, Veselin},
  booktitle={Proceedings of the 2018 conference on empirical methods in natural language processing},
  pages={2475--2485},
  year={2018}
}

@inproceedings{nie2020adversarial,
  title={Adversarial NLI: A new benchmark for natural language understanding},
  author={Nie, Yixin and Williams, Adina and Dinan, Emily and Bansal, Mohit and Weston, Jason and Kiela, Douwe},
  booktitle={Proceedings of the 58th annual meeting of the association for computational linguistics},
  pages={4885--4901},
  year={2020}
}

@inproceedings{mnli,
  title={A broad-coverage challenge corpus for sentence understanding through inference},
  author={Williams, Adina and Nangia, Nikita and Bowman, Samuel},
  booktitle={Proceedings of the 2018 conference of the North American chapter of the association for computational linguistics: human language technologies, volume 1 (long papers)},
  pages={1112--1122},
  year={2018}
}

@inproceedings{snli,
  title={A large annotated corpus for learning natural language inference},
  author={Bowman, Samuel and Angeli, Gabor and Potts, Christopher and Manning, Christopher D},
  booktitle={Proceedings of the 2015 conference on empirical methods in natural language processing},
  pages={632--642},
  year={2015}
}

@article{stokes2020deep,
  title={A deep learning approach to antibiotic discovery},
  author={Stokes, Jonathan M and Yang, Kevin and Swanson, Kyle and Jin, Wengong and Cubillos-Ruiz, Andres and Donghia, Nina M and MacNair, Craig R and French, Shawn and Carfrae, Lindsey A and Bloom-Ackermann, Zohar and others},
  journal={Cell},
  volume={180},
  number={4},
  pages={688--702},
  year={2020},
  publisher={Elsevier}
}

@article{butler2018machine,
  title={Machine learning for molecular and materials science},
  author={Butler, Keith T and Davies, Daniel W and Cartwright, Hugh and Isayev, Olexandr and Walsh, Aron},
  journal={Nature},
  volume={559},
  number={7715},
  pages={547--555},
  year={2018},
  publisher={Nature Publishing Group UK London}
}

@article{Seo2025Paper2CodeACA,
  title={Paper2Code: Automating Code Generation from Scientific Papers in Machine Learning},
  author={Minju Seo and Jinheon Baek and Seongyun Lee and Sung Ju Hwang},
  journal={ArXiv},
  year={2025},
  volume={abs/2504.17192},
  url={https://api.semanticscholar.org/CorpusId:278033490}
}

@article{Kim2025FromRTA,
  title={From Reproduction to Replication: Evaluating Research Agents with Progressive Code Masking},
  author={Gyeongwon James Kim and Alex Wilf and Louis-philippe Morency and Daniel Fried},
  journal={ArXiv},
  year={2025},
  volume={abs/2506.19724},
  url={https://api.semanticscholar.org/CorpusId:280000499}
}

@article{segler2018planning,
  title={Planning chemical syntheses with deep neural networks and symbolic AI},
  author={Segler, Marwin HS and Preuss, Mike and Waller, Mark P},
  journal={Nature},
  volume={555},
  number={7698},
  pages={604--610},
  year={2018},
  publisher={Nature Publishing Group UK London}
}

@article{xie2018crystal,
  title={Crystal graph convolutional neural networks for an accurate and interpretable prediction of material properties},
  author={Xie, Tian and Grossman, Jeffrey C},
  journal={Physical review letters},
  volume={120},
  number={14},
  pages={145301},
  year={2018},
  publisher={APS}
}

@article{You2022ArtificialIIA,
  title={Artificial intelligence in cancer target identification and drug discovery},
  author={Yujie You and Xin Lai and Ying Pan and Huiru Zheng and J. Vera and Suran Liu and Senyi Deng and Le Zhang},
  journal={Signal Transduction and Targeted Therapy},
  year={2022},
  volume={7},
  url={https://doi.org/10.1038/s41392-022-00994-0}
}

@article{Xiang2025SciReplicateBenchBLA,
  title={SciReplicate-Bench: Benchmarking LLMs in Agent-driven Algorithmic Reproduction from Research Papers},
  author={Yanzheng Xiang and Hanqi Yan and Shuyin Ouyang and Lin Gui and Yulan He},
  journal={ArXiv},
  year={2025},
  volume={abs/2504.00255},
  url={https://api.semanticscholar.org/CorpusId:277467991}
}

@article{Vian2024IntegratingAIA,
  title={Integrating Artificial Intelligence for Drug Discovery in the Context of Revolutionizing Drug Delivery},
  author={A. Vișan and I. Neguț},
  journal={Life},
  year={2024},
  volume={14},
  url={https://api.semanticscholar.org/CorpusId:267570328}
}

@article{Siegel2024COREBenchFTA,
  title={CORE-Bench: Fostering the Credibility of Published Research Through a Computational Reproducibility Agent Benchmark},
  author={Zachary S. Siegel and Sayash Kapoor and Nitya Nagdir and Benedikt Stroebl and Arvind Narayanan},
  journal={Trans. Mach. Learn. Res.},
  year={2024},
  volume={2024},
  url={https://arxiv.org/pdf/2409.11363.pdf}
}

@article{Serrano2024ArtificialIA,
  title={Artificial Intelligence (AI) Applications in Drug Discovery and Drug Delivery: Revolutionizing Personalized Medicine},
  author={D. Serrano and F. C. Luciano and B. J. Anaya and Baris Ongoren and Aytug Kara and Gracia Molina and Bianca I Ramirez and Sergio A S{\'a}nchez-Guirales and Jesus A. Simon and Greta Tomietto and Chrysi Rapti and Helga K. Ruiz and Satyavati Rawat and Dinesh Kumar and A. Lalatsa},
  journal={Pharmaceutics},
  year={2024},
  volume={16},
  url={https://api.semanticscholar.org/CorpusId:273366814}
}

@article{Lu2024TheASA,
  title={The AI Scientist: Towards Fully Automated Open-Ended Scientific Discovery},
  author={Chris Lu and Cong Lu and R. T. Lange and J. Foerster and Jeff Clune and David Ha},
  journal={ArXiv},
  year={2024},
  volume={abs/2408.06292},
  url={https://api.semanticscholar.org/CorpusId:271854887}
}

@article{starace2025paperbench,
  title={PaperBench: Evaluating AI's Ability to Replicate AI Research},
  author={Starace, Giulio and Jaffe, Oliver and Sherburn, Dane and Aung, James and Chan, Jun Shern and Maksin, Leon and Dias, Rachel and Mays, Evan and Kinsella, Benjamin and Thompson, Wyatt and others},
  journal={arXiv preprint arXiv:2504.01848},
  year={2025}
}

@article{Jansen2025CodeScientistESA,
  title={CodeScientist: End-to-End Semi-Automated Scientific Discovery with Code-based Experimentation},
  author={Peter Alexander Jansen and Oyvind Tafjord and Marissa Radensky and Pao Siangliulue and Tom Hope and Bhavana Dalvi and Bodhisattwa Prasad Majumder and D. S. Weld and Peter Clark},
  journal={ArXiv},
  year={2025},
  volume={abs/2503.22708},
  url={https://api.semanticscholar.org/CorpusId:277451644}
}

@article{Jansen2024DISCOVERYWORLDAVA,
  title={DISCOVERYWORLD: A Virtual Environment for Developing and Evaluating Automated Scientific Discovery Agents},
  author={Peter Alexander Jansen and Marc-Alexandre C{\^o}t{\'e} and Tushar Khot and Erin Bransom and Bhavana Dalvi and Bodhisattwa Prasad Majumder and Oyvind Tafjord and Peter Clark},
  journal={ArXiv},
  year={2024},
  volume={abs/2406.06769},
  url={https://api.semanticscholar.org/CorpusId:270380311}
}

@article{Yang2024NavigatingDDA,
  title={Navigating Dataset Documentations in AI: A Large-Scale Analysis of Dataset Cards on Hugging Face},
  author={Xinyu Yang and Weixin Liang and James Zou},
  journal={ArXiv},
  year={2024},
  volume={abs/2401.13822},
  url={https://api.semanticscholar.org/CorpusId:267212153}
}

@article{Laufer2025AnatomyOAA,
  title={Anatomy of a Machine Learning Ecosystem: 2 Million Models on Hugging Face},
  author={Benjamin Laufer and Hamidah Oderinwale and Jon Kleinberg},
  journal={ArXiv},
  year={2025},
  volume={abs/2508.06811},
  url={https://api.semanticscholar.org/CorpusId:280566415}
}

@article{Castao2023AnalyzingTEA,
  title={Analyzing the Evolution and Maintenance of ML Models on Hugging Face},
  author={Joel Casta{\~n}o and Silverio Mart{\'i}nez-Fern{\'a}ndez and Xavier Franch and Justus Bogner},
  journal={2024 IEEE/ACM 21st International Conference on Mining Software Repositories (MSR)},
  year={2023},
  pages={607-618},
  url={https://api.semanticscholar.org/CorpusId:265351447}
}

@article{Rahman2025HuggingGraphUTA,
  title={HuggingGraph: Understanding the Supply Chain of LLM Ecosystem},
  author={Mohammad Shahedur Rahman and Peng Gao and Yuede Ji},
  journal={ArXiv},
  year={2025},
  volume={abs/2507.14240},
  url={https://api.semanticscholar.org/CorpusId:280270972}
}

@article{Gao2023OnTOA,
  title={On the Origin of LLMs: An Evolutionary Tree and Graph for 15, 821 Large Language Models},
  author={Sarah Gao and Andrew Gao},
  journal={ArXiv},
  year={2023},
  volume={abs/2307.09793},
  url={https://api.semanticscholar.org/CorpusId:259983028}
}

@article{Silva2025ResearchKGA,
  title={Research Knowledge Graphs in NFDI4DataScience: Key Activities, Achievements, and Future Directions},
  author={Kanishka Silva and Marcel R. Ackermann and Heike Fliegl and G. Gesese and Fidan Limani and Philipp Mayr and Peter Mutschke and A. Oelen and Muhammad Asif Suryani and Sharmila Upadhyaya and Benjamin Zapilko and Harald Sack and Stefan Dietze},
  journal={ArXiv},
  year={2025},
  volume={abs/2508.02300},
  url={https://api.semanticscholar.org/CorpusId:280421789}
}

@article{Shen2023TaskBenchBLA,
  title={TaskBench: Benchmarking Large Language Models for Task Automation},
  author={Yongliang Shen and Kaitao Song and Xu Tan and Wenqi Zhang and Kan Ren and Siyu Yuan and Weiming Lu and Dongsheng Li and Y. Zhuang},
  journal={ArXiv},
  year={2023},
  volume={abs/2311.18760},
  url={https://api.semanticscholar.org/CorpusId:265506220}
}

@inproceedings{Chen2025BenchmarkingRCA,
  title={Benchmarking Recommendation, Classification, and Tracing Based on Hugging Face Knowledge Graph},
  author={Qiaosheng Chen and Kaijia Huang and Xiaofang Zhou and Weiqing Luo and Yuanning Cui and Gong Cheng},
  booktitle={Annual International ACM SIGIR Conference on Research and Development in Information Retrieval},
  year={2025},
  url={https://api.semanticscholar.org/CorpusId:278886655}
}

@inproceedings{Urbanowicz2022STREAMLINEASA,
  title={STREAMLINE: A Simple, Transparent, End-To-End Automated Machine Learning Pipeline Facilitating Data Analysis and Algorithm Comparison},
  author={R. Urbanowicz and Robert F. Zhang and Yuhan Cui and Pranshu Suri},
  booktitle={Genetic Programming Theory and Practice},
  year={2022},
  url={https://api.semanticscholar.org/CorpusId:250048789}
}

@inproceedings{Beel2025EvaluatingSAA,
  title={Evaluating Sakana's AI Scientist: Bold Claims, Mixed Results, and a Promising Future?},
  author={Beel, Joeran and Kan, Min-Yen and Baumgart, Moritz},
  booktitle={ACM SIGIR Forum},
  volume={59},
  pages={1--20},
  year={2025},
  organization={ACM New York, NY, USA}
}

@article{bogin2024super,
  title={Super: Evaluating agents on setting up and executing tasks from research repositories},
  author={Bogin, Ben and Yang, Kejuan and Gupta, Shashank and Richardson, Kyle and Bransom, Erin and Clark, Peter and Sabharwal, Ashish and Khot, Tushar},
  journal={Proceedings of EMNLP},
  year={2024}
}

@article{cheng2025language,
  title={Language Modeling by Language Models},
  author={Cheng, Junyan and Clark, Peter and Richardson, Kyle},
  journal={Proceedings of NeurIPS},
  year={2025}
}

@article{Ait2023OnTSA,
  title={On the Suitability of Hugging Face Hub for Empirical Studies},
  author={Adem Ait and Javier Luis C{\'a}novas Izquierdo and Jordi Cabot},
  journal={ArXiv},
  year={2023},
  volume={abs/2307.14841},
  url={https://api.semanticscholar.org/CorpusId:260203268}
}

@article{bragg2025astabench,
  title={Astabench: Rigorous benchmarking of ai agents with a scientific research suite},
  author={Bragg, Jonathan and D'Arcy, Mike and Balepur, Nishant and Bareket, Dan and Dalvi, Bhavana and Feldman, Sergey and Haddad, Dany and Hwang, Jena D and Jansen, Peter and Kishore, Varsha and Richardson, Kyle and Singh, Amanpreet and Suarana, Harshit and Tiktinsky, Aryen and Vasu, Rosni and Wiener, Guy and Anastasiades, Chloe},
  journal={Proceedings of ICLR},
  year={2026}
}

@article{Castao2024HowDMA,
  title={How do Machine Learning Models Change?},
  author={Joel Casta{\~n}o and Rafael Caba{\~n}as and Antonio Salmer'on and David Lo and Silverio Mart'inez-Fern'andez},
  journal={ArXiv},
  year={2024},
  volume={abs/2411.09645},
  url={https://api.semanticscholar.org/CorpusId:274023512}
}

@article{Lissa2020WORCSAWA,
  title={WORCS: A workflow for open reproducible code in science},
  author={C. J. Lissa and A. Brandmaier and Loek Brinkman and Anna-Lena Lamprecht and Aaron Peikert and Marijn E. Struiksma and Barbara M. I. Vreede},
  journal={Data Sci.},
  year={2020},
  volume={4},
  pages={29-49},
  url={https://api.semanticscholar.org/CorpusId:234357246}
}

@article{yun2021neo,
  title={Neo-gnns: Neighborhood overlap-aware graph neural networks for link prediction},
  author={Yun, Seongjun and Kim, Seoyoon and Lee, Junhyun and Kang, Jaewoo and Kim, Hyunwoo J},
  journal={Advances in Neural Information Processing Systems},
  volume={34},
  pages={13683--13694},
  year={2021}
}

@misc{openai2025gpt52,
  author       = {{OpenAI}},
  title        = {Introducing {OpenAI} GPT-5.2},
  year         = {2025},
  month        = apr,
  url          = {https://openai.com/index/introducing-gpt-5-2/},
  note         = {Accessed: 2026-01-06}
}

@inproceedings{wang2024executable,
  title={Executable code actions elicit better llm agents},
  author={Wang, Xingyao and Chen, Yangyi and Yuan, Lifan and Zhang, Yizhe and Li, Yunzhu and Peng, Hao and Ji, Heng},
  booktitle={Forty-first International Conference on Machine Learning},
  year={2024}
}

@article{Mari2023APWA,
  title={A pragmatic workflow for research software engineering in computational science},
  author={T. Mari{\'c} and Dennis Gl{\"a}ser and Jan-Patrick Lehr and Ioannis Papagiannidis and B. Lambie and Christian H. Bischof and Dieter Bothe},
  journal={ArXiv},
  year={2023},
  volume={abs/2310.00960},
  url={https://api.semanticscholar.org/CorpusId:263605602}
}

@article{Johnson2019ArtifactBasedRHA,
  title={Artifact-Based Rendering: Harnessing Natural and Traditional Visual Media for More Expressive and Engaging 3D Visualizations},
  author={Seth Johnson and F. Samsel and G. Abram and Daniel L. Olson and Andrew J. Solis and Bridger Herman and P. Wolfram and C. Lenglet and Daniel F. Keefe},
  journal={IEEE Transactions on Visualization and Computer Graphics},
  year={2019},
  volume={26},
  pages={492-502},
  url={https://api.semanticscholar.org/CorpusId:199001020}
}

@article{Heumller2020PublishOPA,
  title={Publish or perish, but do not forget your software artifacts},
  author={R. Heum{\"u}ller and Sebastian Nielebock and J. Kr{\"u}ger and F. Ortmeier},
  journal={Empirical Software Engineering},
  year={2020},
  volume={25},
  pages={4585 - 4616},
  url={https://api.semanticscholar.org/CorpusId:220070385}
}

@article{Cooper2022ASRA,
  title={A Systematic Review and Thematic Analysis of Community-Collaborative Approaches to Computing Research},
  author={Ned Cooper and Tiffanie N. Horne and Gillian R. Hayes and Courtney Heldreth and Michal Lahav and Jess Holbrook and Lauren Wilcox},
  journal={Proceedings of the 2022 CHI Conference on Human Factors in Computing Systems},
  year={2022},
  url={https://api.semanticscholar.org/CorpusId:248419416}
}
\bibliographystyle{colm2026_conference}

\clearpage
\appendix

\section{Visualization of Artifact Graph}

\begin{figure}[h]
    \centering
    \includegraphics[width=\linewidth]{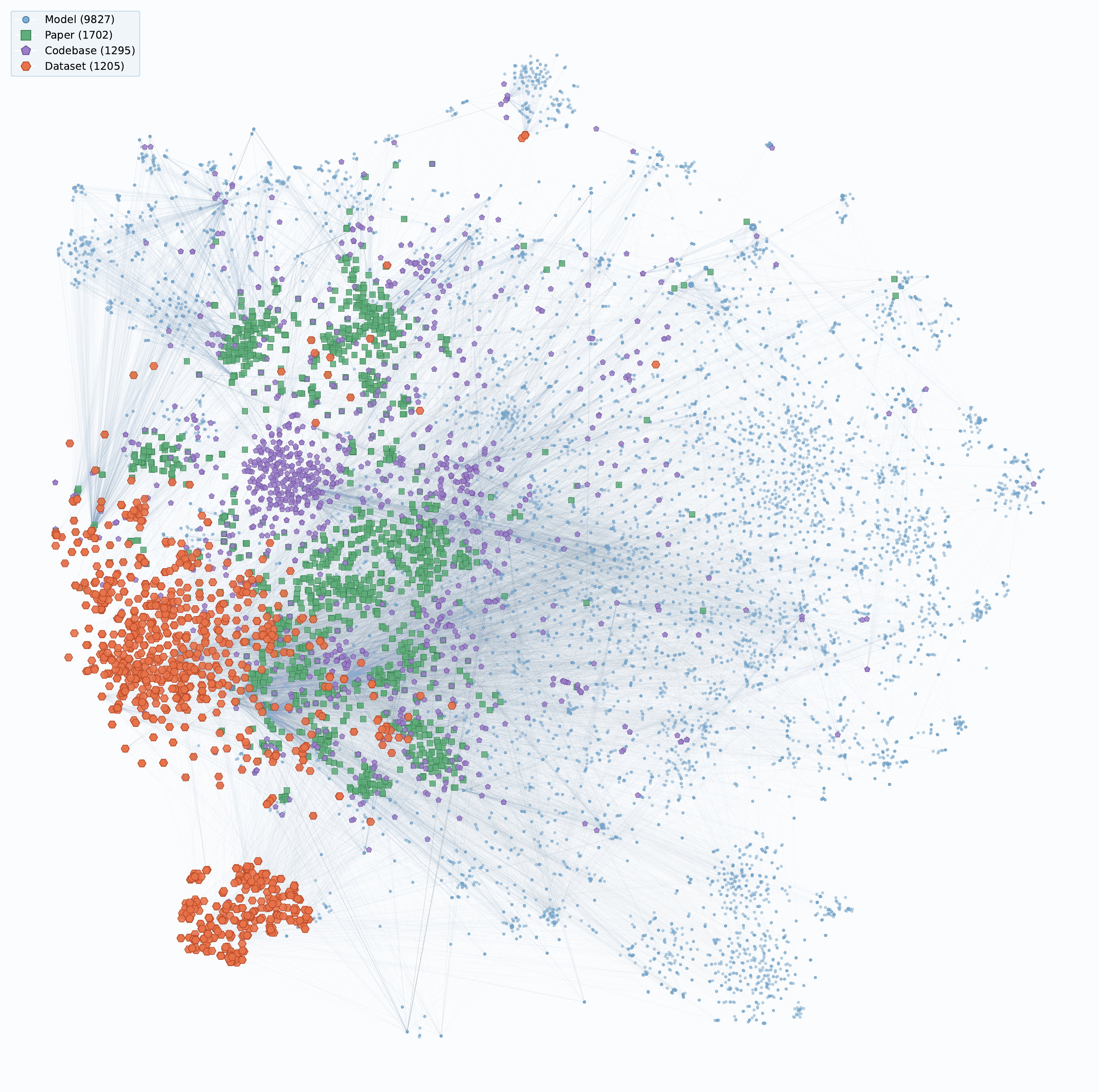}
    \caption{\textbf{Visualization of collected artifact graph}.}
    \label{fig:artifact-graph-full}
\end{figure}

Figure~\ref{fig:artifact-graph-full} shows the full-size visualization of all nodes and edges we included in our collected artifact graph.

\section{Limitations}

\paragraph{Computational considerations in verification} While our two-stage framework effectively reduces the search space through candidate filtering in the prediction phase, the verification stage requires actual code execution for validation. For certain large-scale experiments or resource-intensive datasets, full reproduction may require non-trivial computational costs depending on available hardware resources. We note that our current implementation successfully handles the majority of common datasets, though scaling to extremely high-throughput scenarios across diverse hardware environments remains an interesting direction for future optimization.

\paragraph{Evaluation metric coverage} Our framework currently emphasizes objective, quantitative metrics such as Accuracy, F1, and Exact Match, which are widely reported in model documentation and amenable to automatic extraction. These metrics cover a substantial portion of commonly used evaluation schemes in NLP. While our approach could potentially be extended to incorporate tasks involving human evaluation or qualitative assessment, we leave the integration of such subjective metrics as a natural extension for future work, as they would require additional methodological considerations for automated processing.

\section{Ethics Consideration}
Our work focuses on building an automatic discovery framework over open-source artifacts available on HuggingFace. All experiments are conducted exclusively on publicly released models and datasets that are freely accessible to the research community. We do not introduce any new human or sensitive data, nor do we attempt to deanonymize or misuse existing artifacts. The goal of our framework is to advance automated, reproducible, and scalable scientific discovery, and to help researchers more efficiently identify promising model--dataset interactions. Nevertheless, we acknowledge that automated benchmarking may propagate existing biases and limitations present in the underlying models and datasets. To mitigate this, we emphasize transparency in data collection and reproducibility in our verification pipeline. We encourage the community to view our work as a step toward building more reliable and responsible auto-discovery systems, rather than a replacement for human oversight.

\section{Potential Risks}
\paragraph{Metadata quality considerations} Our prediction stage uses metadata from the HuggingFace community, including model descriptions, architectural specifications, and training configurations. As with any crowdsourced platform, there may occasionally be instances of incomplete or imprecise documentation that could affect initial ranking predictions. 
\paragraph{Performance-oriented discovery scope} \modelname is designed to identify high-performing model-dataset combinations based on standard evaluation metrics. As with any automated performance discovery tool, users should exercise standard research practices by conducting appropriate safety and ethical assessments before deploying discovered models in production environments, particularly for applications involving user-facing content generation. We view our framework as a research tool that augments human decision-making rather than replacing it. Users retain full responsibility for evaluating whether discovered models meet the safety, fairness, and ethical requirements of their specific use cases, and we encourage comprehensive evaluation beyond performance metrics alone.

\section{Scientific Artifacts}

\subsection{Data License}
All data in \benchname are derived from publicly available resources on the HuggingFace Hub, including models, datasets, papers, codebases, and their associated metadata. These resources are released under heterogeneous public licenses, such as Apache 2.0, MIT, and CC-BY. We release \benchname under the Open Database License (ODbL), which permits use, sharing, and modification of the database, while requiring proper attribution and that derivative databases be released under the same license. We will retain license metadata for the original artifacts whenever available and encourage users to comply with the licenses of the underlying resources.

\subsection{Model License}
Our work uses several foundation models for inference, including \texttt{chatgpt-4o-latest}, \texttt{o3-2025-04-16}, \texttt{voyage-3}, and Qwen3-8B. We access \texttt{chatgpt-4o-latest} and \texttt{o3-2025-04-16} through the official OpenAI API, and we use \texttt{voyage-3} through the official VoyageAI inference API. These API-based models are closed-source and governed by their respective proprietary terms of use. We use them only for academic and non-commercial research purposes, with inputs derived from publicly available data. We do not modify these models and use them as provided through their public APIs. Qwen3-8B is released under the Apache 2.0 license and is used for inference in accordance with its license terms.

\subsection{Data Usage}
The data used in this study consist of publicly available scientific artifact information collected from the HuggingFace Hub, including model specifications, dataset descriptions, paper links, codebase metadata, and evaluation results. Our collection focuses exclusively on scientific artifacts and does not intentionally include personally identifiable information. All resources are collected from publicly accessible pages and used for research purposes in accordance with the HuggingFace Hub Terms of Service and the licenses associated with the original artifacts.

\section{Artifact Graph Data Details}
\label{app:artifact-graph}

This section provides additional details on the construction of our artifact
graph and evaluates whether textual artifact metadata introduces shortcut
signals into downstream prediction.

\subsection{Graph Construction Pipeline}
\label{app:graph-construction}

\paragraph{HuggingFace artifact crawling.}
We begin from models and datasets on the HuggingFace Hub and collect their associated metadata and documentation, including model cards, dataset cards, repository identifiers, tags, and referenced artifacts. We retain only models and datasets with at least 1,000 monthly downloads, as low-usage artifacts often provide sparse or insufficiently validated evaluation evidence, making reported results more difficult to reliably verify or reproduce. We retain canonical HuggingFace identifiers whenever available so that each artifact can be mapped back to its original Hub resource. This filtering focuses the graph on widely used and sufficiently documented artifacts while reducing noise from incomplete or low-quality entries.

\paragraph{Evaluation record extraction.}
The primary supervision in our graph comes from evaluation results reported in
HuggingFace model and dataset cards. We parse evaluation tables and textual
result sections into structured records of the form
\[
(m, d, \text{metric}, \text{score}),
\]
where $m$ denotes a model, $d$ a dataset or benchmark, and the remaining fields
specify the reported metric and score. Most evaluation records are directly
extracted from HuggingFace cards. When an entry is incomplete or ambiguous, we
use a search-augmented LLM agent to retrieve supporting evidence from associated
public sources, such as arXiv papers and GitHub repositories, and resolve the
corresponding model, dataset, metric, and score. Multiple metrics for the same
model--dataset pair are retained as separate evaluation records. We normalize
artifact and metric names where possible and discard records that cannot be
reliably resolved.

\paragraph{Contextual enrichment and filtering.}
We further enrich the graph with related arXiv papers, GitHub repositories, and
model--model lineage relations such as fine-tuning. These relations are primarily
obtained from references in artifact cards; when explicit links are unavailable,
the search-augmented LLM agent is used to identify the corresponding public
resources. We remove isolated artifacts (artifact without any edges), entries without usable descriptions,
unresolved references, and low-quality metadata, and merge duplicate references
that resolve to the same artifact. After filtering, the final graph contains
$14{,}053$ nodes and $51{,}337$ edges. A breakdown by node and edge type is shown
in Figure~\ref{fig:artifact-graph-vis}.

\paragraph{Metadata processing.}
The artifact graph requires more semantic information for link prediction and verification. We use an LLM to summarize extracted model and dataset cards, paper abstracts, and
repository descriptions into concise textual node representations. These
representations are used to initialize GNN node features and as textual context
for LLM-based baselines. All quantitative evaluation information, including
metric values and benchmark scores, is removed before constructing these
representations to prevent direct target leakage.

\paragraph{Construction quality audit.}
To assess the reliability of the graph construction pipeline, we randomly sample
100 model--dataset evaluation edges and manually verify each extracted
$(\text{model}, \text{dataset}, \text{metric}, \text{score})$ record against
publicly available evidence. We additionally check whether the reported
evaluation protocol is consistent with the recorded result. Among the sampled
edges, 96 are verified, with the recorded score matching the supporting evidence
within a $2\%$ relative tolerance, while the remaining 4 contain extraction or
protocol inconsistencies.

\subsection{Metadata Leakage Analysis}
\label{app:leakage-analysis}

\begin{table}[t]
\centering
\small
\setlength{\tabcolsep}{12pt}
\begin{tabular}{@{}lcccccccc@{}}
\toprule
& \multicolumn{4}{c}{\textbf{AP}}
& \multicolumn{4}{c}{\textbf{MCC}} \\
\cmidrule(lr){2-5}
\cmidrule(lr){6-9}
\textbf{Backbone}
& \textbf{L0} & \textbf{L1} & \textbf{L2} & \textbf{L3}
& \textbf{L0} & \textbf{L1} & \textbf{L2} & \textbf{L3} \\
\midrule
BUDDY  & 0.215 & 0.223 & 0.208 & 0.226 & 0.234 & 0.226 & 0.228 & 0.234 \\
NeoGNN & 0.210 & 0.227 & 0.208 & 0.209 & 0.208 & 0.202 & 0.210 & 0.195 \\
NCN    & 0.283 & 0.292 & 0.304 & 0.300 & 0.214 & 0.214 & 0.201 & 0.216 \\
NCNC   & 0.260 & 0.269 & 0.289 & 0.275 & 0.248 & 0.256 & 0.248 & 0.251 \\
GATv2  & 0.149 & 0.142 & 0.149 & 0.146 & 0.247 & 0.254 & 0.252 & 0.252 \\
\bottomrule
\end{tabular}
\caption{\textbf{Metadata masking analysis for link prediction.}
L0--L3 progressively remove potential textual shortcut signals.
Performance remains stable across masking levels. We use 3-layer GATv2 as the backbone.}
\label{tab:masking-link}
\end{table}

\begin{table}[t]
\centering
\small
\setlength{\tabcolsep}{12pt}
\begin{tabular}{@{}lcccccccc@{}}
\toprule
& \multicolumn{4}{c}{\textbf{$R^2$}}
& \multicolumn{4}{c}{\textbf{Hits@1}} \\
\cmidrule(lr){2-5}
\cmidrule(lr){6-9}
\textbf{Backbone}
& \textbf{L0} & \textbf{L1} & \textbf{L2} & \textbf{L3}
& \textbf{L0} & \textbf{L1} & \textbf{L2} & \textbf{L3} \\
\midrule
BUDDY  & 0.799 & 0.797 & 0.799 & 0.800 & 0.573 & 0.533 & 0.596 & 0.547 \\
NeoGNN & 0.794 & 0.790 & 0.793 & 0.801 & 0.569 & 0.573 & 0.556 & 0.547 \\
NCN    & 0.788 & 0.797 & 0.798 & 0.798 & 0.533 & 0.578 & 0.560 & 0.542 \\
NCNC   & 0.793 & 0.799 & 0.794 & 0.794 & 0.578 & 0.560 & 0.569 & 0.556 \\
GATv2  & 0.796 & 0.797 & 0.798 & 0.793 & 0.578 & 0.600 & 0.569 & 0.596 \\
\bottomrule
\end{tabular}
\caption{\textbf{Metadata masking analysis for attribute prediction.}
Both $R^2$ and Hits@1 remain stable under progressively stronger masking. We use 3-layer GATv2 as the backbone.}
\label{tab:masking-attr}
\end{table}

\paragraph{Leakage level settings.} Removing numerical evaluation scores alone may not eliminate all potential
shortcut signals. Artifact metadata may still reveal dataset names, task names,
training data, node identities, or qualitative leaderboard claims correlated
with evaluation links. We therefore conduct a masking analysis with four
increasingly aggressive metadata levels:
\textbf{L0}, score-redacted metadata;
\textbf{L1}, additionally masking dataset names;
\textbf{L2}, additionally masking task-family information; and
\textbf{L3}, additionally masking node-identifying entities and
leaderboard-style phrases.

\paragraph{Leakage evaluation.} Table~\ref{tab:masking-link} and \ref{tab:masking-attr} reports link-prediction performance.
Across all GNN backbones, AP and MCC remain largely stable from L0 to L3,
showing no systematic degradation under increasingly aggressive masking. A similar pattern holds for attribute prediction
(Table~\ref{tab:masking-attr}). Both $R^2$ and Hits@1 remain stable across
masking levels, suggesting that score prediction is likewise not driven
primarily by directly identifiable benchmark metadata. Text-based LLM baselines are more sensitive to masking: link-prediction AP
decreases from $0.820$ under L1 to $0.738$ under L3, while MCC decreases from
$0.615$ to $0.419$. This confirms that textual metadata can contain shortcut
signals. In contrast, the stability of the GNN results indicates that our main
graph-based conclusions are not primarily driven by benchmark-name,
node-identity, or leaderboard leakage.

\section{Experimental Details}
\label{exp-details}
\subsection{GNN Architecture Details}
\label{app:gnn-arch}

\modelname{} uses a common encoder--decoder architecture across all GNN
variants. Specifically, every variant shares the same GATv2 graph encoder and
attribute decoder, while differing in the link decoder used to score candidate
model--dataset pairs. The default GATv2 variant uses a bilinear decoder; NCN
and NCNC replace it with their corresponding common-neighbor-based decoders.
For NeoGNN and BUDDY, we implement structure-aware decoders inspired by their
respective structural modeling strategies, rather than reproducing the full
original architectures. The attribute prediction head is shared across all
variants and operates on the same node representations produced by the GATv2
encoder.

\paragraph{Shared GATv2 encoder.}
The encoder stacks three GATv2 message-passing layers, each followed by
\texttt{GraphNorm}, a \texttt{PReLU} activation, and feature dropout
($p{=}0.2$); residual connections are added whenever the input and output
dimensions align. Representations from different depths are aggregated using
JumpingKnowledge (concatenation followed by a linear projection). The first
GATv2 layer projects the $1024$-dimensional Voyage text embeddings to
$128\times 8$ attention heads, intermediate layers preserve the multi-head
representation, and the final layer produces a $128$-dimensional node
embedding. Unless noted otherwise, all variants therefore use the same
single-layer encoder, hidden width $128$, $8$ attention heads, and
$1024$-dimensional input features.

\paragraph{GATv2 (GATv2 + bilinear decoder).}
Our default GATv2 variant uses a bilinear link decoder applied to the two
encoded node representations. Together with the shared regression attribute
head, this model contains approximately $4.80$M parameters.

\paragraph{NCN (GATv2 + NCN decoder).}
The NCN variant keeps the GATv2 encoder unchanged and replaces the bilinear
link decoder with a Neural Common Neighbor (NCN) decoder. For each candidate
model--dataset pair, the decoder incorporates the pooled representations of
their common neighbors together with the pairwise node representations,
providing an explicit first-order structural signal. This variant contains
approximately $4.87$M parameters.

\paragraph{NCNC (GATv2 + NCNC decoder).}
NCNC uses the same GATv2 encoder and extends the NCN link decoder with a
completion module that estimates virtual common-neighbor information for
candidate pairs with sparse observed neighborhoods. The additional completion
module increases the model size to approximately $4.93$M parameters.

\paragraph{NeoGNN-style (GATv2 + structural-overlap decoder).}
Rather than reproducing the full NeoGNN architecture~\citep{yun2021neo}, we adopt its core idea
of augmenting learned node representations with explicit neighborhood-overlap
signals. Our NeoGNN-style decoder incorporates structural features including
common neighbors, Adamic--Adar, and resource allocation statistics, which are
fused with the GATv2 node representations before link scoring. This variant
contains approximately $4.87$M parameters.

\paragraph{BUDDY-style (GATv2 + sketch-based decoder).}
Similarly, our BUDDY-style variant does not reproduce the complete original
BUDDY architecture~\citep{chamberlain2022graph}. Instead, it adopts its sketch-based structural modeling
principle by augmenting the GATv2 representations with compact neighborhood
sketch features that approximate local subgraph structure without explicit
subgraph extraction for each candidate pair. This model contains approximately
$4.87$M parameters.

\paragraph{Joint training.}
All variants share the same attribute regression decoder and are trained
jointly using the link and attribute objectives, with
$\lambda_{\mathrm{attr}}=5$. Consequently, the comparison primarily isolates
the effect of different link-side structural inductive biases while keeping the
underlying graph encoder and attribute predictor fixed. We refer to the latter
two variants as NeoGNN-style and BUDDY-style to emphasize that they instantiate
the corresponding structural ideas within our shared GATv2 encoder--decoder
framework rather than serving as exact reproductions of the original models.

\subsection{Training Details}
\label{app:joint-training}
Our primary \modelname{} experiments jointly train both decoder heads on a
single shared encoder; the link and attribute objectives are optimized
together so the encoder learns representations useful for both tasks.

\paragraph{Negative sampling.} Positive examples are the observed
(model, dataset, metric) triples in the training split. For every positive we
sample $2$ negative (model, dataset) pairs uniformly from pairs absent in all
splits (negative sampling ratio $2$). Negatives enter only the link term; the
attribute term is computed on positive edges alone.

\paragraph{Training objective.} We minimize
$\mathcal{L} = \mathcal{L}_{\text{link}} + 5.0\cdot
\mathcal{L}_{\text{attr}}$. The link term $\mathcal{L}_{\text{link}}$ is a
class-balanced binary cross-entropy: the mean of a positive-only and a
negative-only BCE, so positives and negatives contribute equally regardless of
the $1{:}2$ sampling ratio. The attribute term $\mathcal{L}_{\text{attr}}$ is
a mean-squared error on positive edges. The attribute head is
\emph{link-conditioned}: it receives the predicted link logit together with
the pooled pair representation, rather than scoring the pair in isolation.

\paragraph{Attribute target.}
For attribute prediction, we use mean squared error (MSE) as the regression loss. Targets $y\in[0,1]$ are clamped to $[10^{-7},,1-10^{-7}]$ and mapped to logit space via
$\mathrm{logit}(y)=\log\frac{y}{1-y}$. We then minimize the MSE between the predicted raw attribute logits and the logit-transformed targets. At evaluation, the predicted logits are clipped to $[-10,10]$ before applying a sigmoid to recover metric values in $[0,1]$.

\paragraph{Optimization.} We use \textbf{Adam}~\citep{kingma2014adam} with
learning rate $2\times10^{-3}$ and weight decay $10^{-5}$, under a
cosine-annealing schedule that decays the learning rate to $10^{-5}$ over
$1{,}500$ epochs.

\paragraph{Model selection.}
Because the available training data size is limited, we do not further partition the training data to construct a separate validation set. Instead, all models are trained for a fixed budget of 1,500 optimization steps, and we report the checkpoint from the final training step without any form of model selection. The same checkpoint is used across all downstream tasks and evaluation metrics on test-set performance. All methods are evaluated using the same splits and candidate construction,
so the GNN, heuristic, reranker, and LLM-based methods are directly comparable
within each task.

\paragraph{Link task evaluation data.}
For link prediction and link ranking, we evaluate model--dataset edges under
two settings. The \emph{transductive} setting uses an edge-level split with
$\text{test\_ratio}{=}0.2$ and seed $42$. The \emph{inductive} setting uses a
disjoint model partition, where test models are unseen during training. In both
settings, evaluation negatives are constructed by full enumeration rather than
subsampling of models: any model--dataset pair that is not an observed positive in any
split is treated as a negative candidate. This yields approximately 5.3M
evaluation pairs in the transductive link-prediction setting, with positive
prevalence around 0.1\%.

\paragraph{Attribute task evaluation data.}
Attribute prediction and attribute ranking are evaluated only on test positive
edges with valid numeric metric metadata. Ground-truth values are obtained from
the normalized edge metadata, where all numeric targets are scaled to $[0,1]$.
The scalar target is selected differently for the two attribute tasks. For
attribute prediction, we select one metric per edge: the alphabetically first
numeric metric available for that edge. Edges without any numeric metric are
excluded. For attribute ranking, we select one metric per dataset: the most
frequent numeric metric among that dataset's test edges. Only edges containing
the selected metric are retained, and datasets are excluded if they have fewer
than two valid edges or if all target values are identical. Therefore,
attribute prediction and attribute ranking are evaluated on related but not
identical subsets of test edges.

\paragraph{Link verification data.}
For link verification, we construct a curated benchmark of 263 model--dataset pairs with sufficiently well-specified evaluation protocols and computationally feasible artifacts. We acknowledge that this selection introduces a bias toward popular and easier-to-execute artifacts; however, the purpose of this benchmark is to enable controlled comparison of verification agents rather than to estimate success over arbitrary real-world artifacts. Importantly, even under this relatively favorable setting, automatic reproduction remains challenging: our full agent reproduces only $72.6\%$ of the pairs, with substantially lower success under ablations. We further evaluate robustness in a less curated setting by randomly sampling 50 model--dataset edges without the feasibility filters. In this in-the-wild stress test, only 5 of 50 pairs execute successfully, and only 1 reproduces the reported score within $20\%$ relative error. Common failures arise from quantized or very large models, nonstandard loading procedures, and incomplete evaluation documentation. Together, these results suggest that the curated 263-pair benchmark already presents a nontrivial verification challenge, while fully automatic verification of arbitrary artifacts remains substantially more difficult.

\paragraph{Link prediction setting.}
Link prediction is evaluated over the full pool of test positives and fully
enumerated negatives. This setting measures whether a method can distinguish
observed model--dataset links from all unobserved pairs under extreme class
imbalance. MCC is computed using a fixed decision threshold; the joint GNN uses
a sigmoid threshold of $0.5$, while heuristic baselines use $0.9$.

\paragraph{Link ranking setting.}
Link ranking is evaluated independently for each test dataset. The candidate
set contains the dataset's positive test models together with all other models
that are not known positives for that dataset across train and test. This
creates approximately $10^4$ candidates per dataset, again with full negatives
and no sampling. This setting measures whether a method can rank true
model--dataset links above all plausible alternatives for the same dataset.

\paragraph{Attribute prediction setting.}
Attribute prediction is evaluated on the per-edge scalar targets. The
model predicts an attribute logit $\hat{\ell}$, which is converted to a bounded
score by
\[
    \hat{y} = \sigma(\mathrm{clip}(\hat{\ell}, -10, 10)).
\]
The prediction is compared against the normalized ground-truth value
$y\in[0,1]$. This setting measures whether a method can estimate the expected
evaluation score of a known model--dataset edge.

\paragraph{Attribute ranking setting.}
Attribute ranking is evaluated at the dataset level using the selected metric
for each qualifying dataset. This setting measures whether a method can rank
models by their relative performance on the same dataset, rather than merely
predicting calibrated scalar scores across heterogeneous metrics. The NDCG@1
reported for this task uses a continuous top-1 regret-ratio form,
\[
    \mathrm{NDCG@1}
    =
    \frac{y_{\text{top-1}}}{\max_i y_i},
\]
which differs from the binary-relevance NDCG used in link ranking. Because this
metric is saturated, simple mean-based baselines can already obtain high
scores.

\paragraph{Link verification data}

\subsection{Compute and Budget}
\label{app:budget}
The GNN-based \modelname{} checkpoint is $\sim$5M parameters (<100\,MB on
disk). GNN training runs in <1 hour per configuration on a single
NVIDIA A100. Node text embeddings are computed once via the Voyage AI
\texttt{voyage-3} endpoint (dimension 1024) and cached; embedding the
14{,}053-node graph is a one-time API charge under \$20. LLM-based
baselines (verification, prediction, ranking) over the full NLI
evaluation suite total <\$1{,}000. Closed-source baseline model sizes
are not publicly disclosed.

\section{Details of Case Study}
\label{appendix:case-study-detail}

\paragraph{Dataset name} We use short names for dataset identifiers in figures and tables.
Table~\ref{tab:dataset-short-names} lists the correspondence between each dataset short name and its full dataset ID.

\paragraph{Model name} To improve readability, we use short names for model identifiers in figures and tables.
Table~\ref{tab:model-short-names} lists the correspondence between each short name and its full HuggingFace model ID.

  \begin{table}[h]
  \centering
  \caption{{Mapping between NLI dataset short names and HuggingFace dataset identifiers.}}
  \label{tab:dataset-short-names}
  \small
  \begin{tabular}{l l}
  \hline
  \textbf{Short name} & \textbf{Full dataset ID} \\
  \hline
  SNLI               & stanfordnlp/snli \\
  MNLI               & nyu-mll/multi\_nli \\
  RTE                & SetFit/rte \\
  QNLI               & SetFit/qnli \\
  ANLI               & facebook/anli \\
  SciTail            & allenai/scitail \\
  WANLI              & alisawuffles/WANLI \\
  MedNLI             & araag2/MedNLI \\
  bAbI-NLI           & tasksource/babi\_nli \\
  $\delta$-NLI       & tasksource/defeasible-nli \\
  NLI-FEVER          & pietrolesci/nli\_fever \\
  RobustNLI-ST-SE    & pietrolesci/robust\_nli (ST\_SE split) \\
  \hline
  \end{tabular}
  \end{table}

  \begin{table*}[t]
  \centering
  \caption{{Mapping between model short names and full HuggingFace model identifiers (all 45 evaluated models).}}
  \label{tab:model-short-names}
  \small
  \begin{tabular}{l l}
  \hline
  \textbf{Short name} & \textbf{Full model ID} \\
  \hline
  \multicolumn{2}{l}{\textit{Zero-shot NLI}} \\
  DeBERTa-L-ZS       & MoritzLaurer/deberta-v3-large-zeroshot-v2.0 \\
  RoBERTa-L-ZS       & MoritzLaurer/roberta-large-zeroshot-v2.0-c \\
  XtremeDistil       & MoritzLaurer/xtremedistil-l6-h256-zeroshot-v1.1-all-33 \\
  mDeBERTa-XNLI      & MoritzLaurer/mDeBERTa-v3-base-xnli-multilingual-nli-2mil7 \\
  DeBERTa-L-MFAW     & MoritzLaurer/DeBERTa-v3-large-mnli-fever-anli-ling-wanli \\
  DeBERTa-B-MFA      & MoritzLaurer/DeBERTa-v3-base-mnli-fever-anli \\
  \hline
  \multicolumn{2}{l}{\textit{Multi-task NLI}} \\
  DeBERTa-L-Task     & sileod/deberta-v3-large-tasksource-nli \\
  DeBERTa-B-Task     & sileod/deberta-v3-base-tasksource-nli \\
  ModernBERT-L       & tasksource/ModernBERT-large-nli \\
  ModernBERT-B       & tasksource/ModernBERT-base-nli \\
  DeBERTa-S-Long     & tasksource/deberta-small-long-nli \\
  FineCat-L          & dleemiller/finecat-nli-l \\
  ModernCE-L         & dleemiller/ModernCE-large-nli \\
  \hline
  \multicolumn{2}{l}{\textit{SNLI+MNLI+FEVER+ANLI}} \\
  ALBERT-XXL-R3      & ynie/albert-xxlarge-v2-snli\_mnli\_fever\_anli\_R1\_R2\_R3-nli \\
  RoBERTa-L-R3       & ynie/roberta-large-snli\_mnli\_fever\_anli\_R1\_R2\_R3-nli \\
  BART-L-R3          & ynie/bart-large-snli\_mnli\_fever\_anli\_R1\_R2\_R3-nli \\
  ELECTRA-L-R3       & ynie/electra-large-discriminator-snli\_mnli\_fever\_anli\_R1\_R2\_R3-nli \\
  DeBERTa-L-R3       & Joelzhang/deberta-v3-large-snli\_mnli\_fever\_anli\_R1\_R2\_R3-nli \\
  DeBERTa-MSA        & NDugar/debertav3-mnli-snli-anli \\
  \hline
  \multicolumn{2}{l}{\textit{Cross-encoder NLI}} \\
  CE-DeBERTa-L       & cross-encoder/nli-deberta-v3-large \\
  CE-DeBERTa-B       & cross-encoder/nli-deberta-v3-base \\
  CE-DeBERTa-S       & cross-encoder/nli-deberta-v3-small \\
  CE-RoBERTa-B       & cross-encoder/nli-roberta-base \\
  \hline
  \multicolumn{2}{l}{\textit{DeBERTa-MNLI}} \\
  DeBERTa-B-MNLI     & microsoft/deberta-base-mnli \\
  DeBERTa-L-MNLI     & microsoft/deberta-large-mnli \\
  DeBERTa-XL-MNLI    & microsoft/deberta-xlarge-mnli \\
  DeBERTa-v2-XL      & microsoft/deberta-v2-xlarge-mnli \\
  DeBERTa-v2-XXL     & microsoft/deberta-v2-xxlarge-mnli \\
  DeBERTa-L-MNLI-K   & khalidalt/DeBERTa-v3-large-mnli \\
  DeBERTa-L-SNLI-P   & pepa/deberta-v3-large-snli \\
  DeBERTa-L-SNLI-U   & utahnlp/snli\_microsoft\_deberta-v3-large\_seed-1 \\
  \hline
  \multicolumn{2}{l}{\textit{BART/RoBERTa}} \\
  BART-L-MNLI        & facebook/bart-large-mnli \\
  BART-Yahoo         & joeddav/bart-large-mnli-yahoo-answers \\
  RoBERTa-L-MNLI     & roberta-large-mnli \\
  RoBERTa-L-WANLI    & alisawuffles/roberta-large-wanli \\
  \hline
  \multicolumn{2}{l}{\textit{Domain-specific / smaller}} \\
  PubMedBERT-NLI     & pritamdeka/PubMedBERT-MNLI-MedNLI \\
  Erlangshen-RoBERTa & IDEA-CCNL/Erlangshen-Roberta-330M-NLI \\
  DistilCamemBERT    & cmarkea/distilcamembert-base-nli \\
  DistilBERT         & typeform/distilbert-base-uncased-mnli \\
  ALBERT-B-MNLI      & prajjwal1/albert-base-v2-mnli \\
  BERT-B-SNLI        & textattack/bert-base-uncased-snli \\
  \hline
  \multicolumn{2}{l}{\textit{Generative LLM}} \\
  GPT-Neo-1.3B       & EleutherAI/gpt-neo-1.3B \\
  Gemma-2B           & google/gemma-2b \\
  Gemma-2-2B         & google/gemma-2-2b \\
  Gemma-3-1B         & google/gemma-3-1b-it \\
  \hline
  \end{tabular}
  \end{table*}

\end{document}